\documentclass{article}

\usepackage{microtype}
\usepackage{graphicx}
\usepackage{subcaption}
\usepackage{booktabs} % for professional tables
\usepackage{xcolor}   % for \textcolor
\usepackage{multirow} % for \multirow

\usepackage{tabularx}
\usepackage{array}
\usepackage{dblfloatfix}
\usepackage{placeins}
\usepackage[most]{tcolorbox} % for tcolorbox
\usepackage{forest}          % for the decision tree
\usepackage{arydshln}

\usepackage{hyperref}

\usepackage[preprint]{icml2026}

\makeatletter
\let\algorithmicrequire\relax
\let\algorithmicensure\relax
\let\algorithmicwhile\relax
\let\algorithmicendwhile\relax
\makeatother
\usepackage{algpseudocode}

\usepackage{amsmath}
\usepackage{amssymb}
\usepackage{mathtools}
\usepackage{amsthm}

\usepackage[capitalize,noabbrev]{cleveref}

\theoremstyle{plain}

\theoremstyle{definition}

\theoremstyle{remark}

\usepackage[disable,textsize=tiny]{todonotes}
\icmltitlerunning{ACT: Agentic Classification Tree}

\begin{document}

\twocolumn[
  \icmltitle{ACT: Agentic Classification Tree}



  \icmlsetsymbol{equal}{*}

    \begin{icmlauthorlist}
      \icmlauthor{Vincent Grari}{axa,trail}
      \icmlauthor{Tim Arni}{axa,epfl}
      \icmlauthor{Thibault Laugel}{axa,trail}
      \icmlauthor{Sylvain Lamprier}{leria}
      \icmlauthor{James Zou}{stanford}
      \icmlauthor{Marcin Detyniecki}{axa,trail,ibspan}
    \end{icmlauthorlist}
    
    \icmlaffiliation{axa}{AXA AI Research}
    \icmlaffiliation{stanford}{Stanford University}
    \icmlaffiliation{epfl}{EPFL, Lausanne, Switzerland}
    \icmlaffiliation{trail}{TRAIL, Sorbonne Universit\'e, Paris, France}
    \icmlaffiliation{leria}{LERIA, Universit\'e d’Angers, France}
    \icmlaffiliation{ibspan}{Polish Academy of Sciences, IBS PAN, Warsaw, Poland}

    \icmlcorrespondingauthor{Vincent Grari}{vincent.grari@axa.com}
    \icmlcorrespondingauthor{Tim Arni}{tim.arni@epfl.ch}

  % \begin{icmlauthorlist}
  %   \icmlauthor{Firstname1 Lastname1}{equal,yyy}
  %   \icmlauthor{Firstname2 Lastname2}{equal,yyy,comp}
  %   \icmlauthor{Firstname3 Lastname3}{comp}
  %   \icmlauthor{Firstname4 Lastname4}{sch}
  %   \icmlauthor{Firstname5 Lastname5}{yyy}
  %   \icmlauthor{Firstname6 Lastname6}{sch,yyy,comp}
  %   \icmlauthor{Firstname7 Lastname7}{comp}
  %   %\icmlauthor{}{sch}
  %   \icmlauthor{Firstname8 Lastname8}{sch}
  %   \icmlauthor{Firstname8 Lastname8}{yyy,comp}
  %   %\icmlauthor{}{sch}
  %   %\icmlauthor{}{sch}
  % \end{icmlauthorlist}



  \icmlkeywords{Machine Learning, ICML
  }

  \vskip 0.3in
]



\printAffiliationsAndNotice{}  % no special notice (required even if empty)

% Added for preprint to keep double blind review
\hypersetup{
  pdfsubject={},
  pdfkeywords={Machine Learning}
}

\begin{abstract}
  When used in high-stakes settings, AI systems are expected to produce decisions that are transparent, interpretable and auditable—a requirement increasingly expected by regulations. Decision trees such as CART provide clear and verifiable rules, but they are restricted to structured tabular data and cannot operate directly on unstructured inputs such as text. In practice, large language models (LLMs) are widely used for such data, yet prompting strategies such as chain-of-thought or prompt optimization still rely on free-form reasoning, limiting their ability to ensure trustworthy behaviors. We present the \textit{Agentic Classification Tree} (ACT), which extends decision-tree methodology to unstructured inputs by formulating each split as a natural-language question, refined through impurity-based evaluation and LLM feedback via TextGrad. Experiments on text benchmarks show that ACT matches or surpasses prompting-based baselines while producing transparent and interpretable decision paths. \textcolor{red}{\textbf{Code:} \url{https://github.com/axa-rev-research/ACT}}
\end{abstract}

\section{Introduction}
\label{sec:introduction}

\begin{figure*}[!tb]
\centering

% ---- Patient Descriptions Box ----
\begin{minipage}{\textwidth}
    \centering
    \begin{tcolorbox}[
        enhanced,
        colback=purple!10,
        colframe=purple!60!black,
        width=\linewidth,
        arc=2mm,
        boxrule=1.2pt,
        boxsep=1pt,
        fonttitle=\bfseries,
        title=Unstructured Data: Tuberculosis diagnosis based on symptoms descriptions,
        sidebyside,
        sidebyside gap=10pt,
        lefthand width=0.9\textwidth,
        left=1pt,
        right=1pt,
        top=1pt,
        bottom=1pt
    ]
    \textbf{Patient 1:} "My main symptom is a runny nose with increased sweating, pain, fever, skin rashes, nasal congestion, sore throat, muscle pain, loss of appetite, and chills. The heavy pain is in my left temple ... " %, and I have pink rashes on the back of my neck." 

    \textbf{Patient 2:} "I am coughing blood with pain (heavy in my left temple), skin lesions, nasal congestion, extreme fatigue affecting activities, diffuse muscle pain, loss of appetite, chills, a heavy pain on my left temple ... " %, and pink rash on the right side of my neck."

    \tcblower
    \begin{tabular}{c}\hspace{-0.2cm}\includegraphics[width=0.6\linewidth]{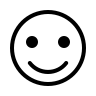}\\
    \hspace{-0.2cm}\includegraphics[width=0.6\linewidth]{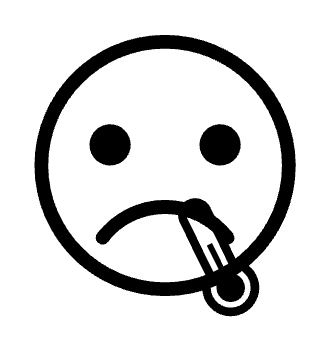}
    \end{tabular}
    \end{tcolorbox}
\end{minipage}

\vspace{0.3cm}

% ---- Decision Tree ----
\begin{minipage}{\textwidth}
    \centering
    \resizebox{0.90\textwidth}{!}{%
    \begin{forest}
    for tree={
        align=center,
        parent anchor=south,
        child anchor=north,
        l sep+=12pt,
        s sep+=6pt,
        draw,
        rounded corners=4pt,
        minimum width=2.4cm,
        minimum height=0.8cm,
        font=\footnotesize,
        edge={->, >=latex, thick},
    }
    [{Does this example involve coughing \\ up blood or weight loss?}, fill=blue!10
        [{Does the example show coughing up blood \\ with fever or intense fatigue?}, edge label={node[midway, font=\scriptsize, fill=white, inner sep=1pt]{yes}}, fill=blue!10
            [{\textbf{TB}}, edge label={node[midway, font=\scriptsize, fill=white, inner sep=1pt]{yes}}, fill=teal!30, draw=teal!70!black, thick]
            [{Does the example show evidence of severe \\ pain and either shortness of breath \\ or swelling in unusual locations?}, edge label={node[midway, font=\scriptsize, fill=white, inner sep=1pt]{no}}, fill=blue!10
                [{\textbf{TB}}, edge label={node[midway, font=\scriptsize, fill=white, inner sep=1pt]{yes}}, fill=teal!30, draw=teal!70!black, thick]
                [{\textbf{Not TB}}, edge label={node[midway, font=\scriptsize, fill=white, inner sep=1pt]{no}}, fill=orange!30, draw=orange!70!black, thick]
            ]
        ]
        [{Does the example show fever \\ and swollen lymph nodes?}, edge label={node[midway, font=\scriptsize, fill=white, inner sep=1pt]{no}}, fill=blue!10
            [{Does the example show itchy or \\runny nose or itchy eyes?}, edge label={node[midway, font=\scriptsize, fill=white, inner sep=1pt]{yes}}, fill=blue!10
                [{\textbf{Not TB}}, edge label={node[midway, font=\scriptsize, fill=white, inner sep=1pt]{yes}}, fill=orange!30, draw=orange!70!black, thick]
                [{\textbf{TB}}, edge label={node[midway, font=\scriptsize, fill=white, inner sep=1pt]{no}}, fill=teal!30, draw=teal!70!black, thick]
            ]
            [{\textbf{Not TB}}, edge label={node[midway, font=\scriptsize, fill=white, inner sep=1pt]{no}}, fill=orange!30, draw=orange!70!black, thick]
        ]
    ]
    \end{forest}
    }
\end{minipage}

% \caption{
% Example ACT decision tree for tuberculosis diagnosis using unstructured, free-text patient descriptions. A tree is automatically learned, with each node containing a binary natural language question, autonomously discovered via recursive prompt refinement to maximize label separation at each split. At inference, these questions are answered by a large language model (LLM) from the root node to the leaves. The final classification (TB or Not TB) corresponds to the majority label of training examples described by each leaf.
% }
\caption{
Example ACT for tuberculosis (TB) diagnosis using unstructured, free-text patient descriptions. A decision tree is automatically learned, with each node containing a binary natural-language question, autonomously discovered via recursive prompt refinement to maximize label separation at each split. At inference, given an input sample, these questions are answered by an LLM from the root node to the leaves. The final classification (TB or Not TB) corresponds to the majority label of training examples described by each leaf.
}
\label{fig:full_combined_tb}
\end{figure*}

As AI systems become increasingly integrated into high-stakes domains such as healthcare, education, legal decision-making and finance, the need for transparency, interpretability and auditability in AI decision-making has intensified. These requirements are grounded not only in practical considerations, but also in legal obligations: recent governance frameworks and regulations such as the EU AI Act %\footnote{https://artificialintelligenceact.eu/}
or %, the OECD AI Principles\footnote{https://oecd.ai/en/ai-principles},
the NIST AI Risk Management Framework %\footnote{https://www.nist.gov/itl/ai-risk-management-framework}
emphasize that AI-based decisions in high-stakes scenarios must be explainable and subject to human oversight. In this context, there is a growing need for models whose decision processes can be inspected, verified and understood not only by technical experts but also by stakeholders, auditors and regulators.

Historically, complex decision-making tasks in high-stakes domains %—such as medical diagnosis or fraud detection %, or credit underwriting—
have been addressed by interpretable AI systems such as expert systems~\citep{jackson1986introduction,shortliffe1986medical,leonard1993detecting,talebzadeh1995countrywide} and decision trees~\citep{DBLP:books/wa/BreimanFOS84}. These models allow users to trace each decision through a sequence of explicit, human-understandable rules~{\color{blue}\cite{slack2019assessing}}, directly supporting the requirements set out by regulatory frameworks\footnote{https://airc.nist.gov/airmf-resources/playbook/}. However, their applicability is largely limited to structured inputs such as tabular data and are not directly applicable to unstructured data such as natural language or images.

Recent advancements in large language models (LLMs) have substantially enhanced the capacity for semantic understanding and reasoning within NLP %natural language processing
~\citep{brown2020language, achiam2023gpt}.
Building on these models, agentic systems~\citep{shinn2023reflexion,yao2023react,schick2023toolformer} have been proposed to address complex tasks involving reasoning over diverse types of data, and are seen as a very promising direction for future AI systems~\citep{chen2023agentverse,wang2024survey}. Yet, despite their impressive capabilities, LLMs still suffer from several drawbacks—such as hallucinations, %inconsistencies,
unreliability and difficulty to audit~\citep{kryscinski2019evaluating, ji2023survey, tamkin2021understanding}. These issues constitute major obstacles to building reliable AI systems, limiting their adoption in sensitive or regulated domains. A growing body of work seeks to address these weaknesses by guiding LLM behavior toward more structured or self-consistent reasoning. Chain-of-Thought prompting~\cite{wei2022chain}, for example, encourages multi-step reasoning% by inserting explicit intermediate steps into prompts,
, improving consistency and sometimes reducing hallucinations. Other techniques, such as self-reflection~\cite{shinn2023reflexion, madaan2023self} and prompt optimization~\cite{zhou2022large,ji2023towards,yuksekgonul2024textgrad,renze2024self}, introduce agentic feedback loops to iteratively refine model outputs. While these methods improve reliability, %in practice,
they still rely on free-form text generation and implicit reasoning %patterns
that remain difficult to verify or audit. %, or formally constrain.

% In this paper, we introduce a new type of classifier designed to combine the semantic reasoning capabilities of LLMs with the transparency of decision trees. The idea is to follow a divide-and-conquer paradigm, allowing to reduce hallucination and reasoning errors by successive decomposition of the problem following a hierarchical decision logic. We therefore propose to structure LLM and agentic reasoning to 
% address complex classification tasks through verifiable decision patterns on \emph{unstructured data} such as images or texts, in the form of an agentic decision tree.

In this paper, we introduce a new type of classifier designed to combine the semantic reasoning capabilities of LLMs with the transparency of decision trees. The idea is to follow a divide-and-conquer paradigm by decomposing a difficult classification question (e.g., \textit{Does the patient have disease X?}) into a series of easier questions (e.g., \textit{Is symptom Y present?}), as shown in Figure~\ref{fig:full_combined_tb}. We therefore propose to structure LLM and agentic reasoning as an agentic decision tree, addressing complex classification tasks on \textit{unstructured data} through verifiable decision patterns while reducing reasoning errors via hierarchical problem decomposition.

% Adapting traditional decision tree algorithms such as CART~\cite{DBLP:books/wa/BreimanFOS84} and C4.5~\cite{quinlan1993c4}, we propose the Agentic Classification Tree (ACT) where each node is defined as a binary natural language question over the input data, iteratively partitioning the data into two subsets at each step. 

Adapting traditional decision tree algorithms such as CART~\cite{DBLP:books/wa/BreimanFOS84} and C4.5~\cite{quinlan1993c4}, we propose the Agentic Classification Tree (ACT). In ACT, each node is defined as a binary natural-language question over the input data, iteratively partitioning the data into two subsets at each step. The best splits are found using TextGrad~\cite{yuksekgonul2024textgrad}, a prompt-refinement technique providing textual feedback to an agent, to minimize the Gini criterion~\cite{DBLP:books/wa/BreimanFOS84} at each split. 
Ultimately, ACT takes the form of a decision tree (cf. Figure~\ref{fig:full_combined_tb}), each leaf leading to a class prediction after successive splitting. The benefits of ACT are therefore the following:
\begin{itemize}

    \item \textbf{Transparent and structured decision process:} ACT enforces a systematic, rule-based structure that aligns the model's internal reasoning with its observable decision process. Its tree-based architecture provides fully traceable and interpretable decision paths, facilitating auditability, human oversight and intervention. %—akin to the properties of traditional expert systems.

    \item \textbf{Optimized decision process:} we show that decomposing some tasks into subquestions improves the performance of LLMs without retraining. % By automatically finding the most relevant questions at each node, ACT is able to surpass existing methods.

\end{itemize}

\section{Problem Setup and Related Work}\label{sec:related_work}

We consider a binary classification problem over an \emph{unstructured} input space $\mathcal{X}$ (e.g., text or images), with labels $y \in \{0,1\}$. Given a dataset $\mathcal{D} = \{(x_i, y_i)\}_{i=1}^N$ sampled i.i.d. from an unknown distribution, the goal is to learn a classifier $f: \mathcal{X} \rightarrow \{0,1\}$ that is both accurate and interpretable.
There is an abundant literature  in XAI that focuses on defining desiderata for interpretable models~\cite{doshi2017towards,lipton2018mythos}.
Building on the regulatory requirements discussed in Section~\ref{sec:introduction}, we therefore aim to build a decision model satisfying the following properties:
\begin{itemize}
    \item \textbf{Accuracy:} competitive predictive performance on unstructured inputs;
    \item \textbf{Transparency:} decisions should be produced via interpretable, step-by-step reasoning; 
    \item \textbf{Contestability:} the ability to contest an algorithm-based decision by disputing the conditions it relies on~\cite{wachter2017counterfactual,venkatasubramanian2020philosophical,lyons2021conceptualising}.
\end{itemize}

Translating these desiderata into technical requirements is challenging given the opacity of modern architectures such as LLMs and VLMs, with limited literature guidance on suitable explanations in these contexts. %Translating these desiderata into technical requirements is made especially difficult given the opaqueness and complexity of modern architectures such as LLMs and VLMs. There is a clear lack of literature support for what explanations satisfying these properties should look like in these contexts.
As a result, in this work, we propose to define interpretability through the notion of \textbf{explicit decision paths}—structured sequences of semantically meaningful reasoning steps that can be inspected, understood and verified by human users. 
Below, we ground this proposition in the existing XAI and LLM literatures.

\paragraph{Interpretable models and post-hoc explanations.}
Interpretable models such as decision trees~\citep{DBLP:books/wa/BreimanFOS84, quinlan1993c4} are traditionally viewed as a natural solution to meet the three aforementioned criteria. %Offering transparent, rule-based decision paths that are easily understood and audited, their hierarchical structure allows users to trace predictions through explicit, human-readable splits, making them well-suited for regulated or high-stakes domains~\citep{rudin2019stop}.
Their transparent, rule-based structure allows users to trace predictions through human-readable splits, making them well-suited for regulated or high-stakes domains~\citep{rudin2019stop}. However, they are fundamentally limited to structured inputs, relying on predefined features and struggling to handle unstructured data such as raw text or images. 
Efforts to address interpretability in unstructured data contexts include post-hoc explanation techniques such as LIME~\citep{ribeiro2016should} and SHAP~\citep{lundberg2017unified} proposing explanations in the form of feature attributions at the pixel (for images) or the word level (for text). Although very popular, these methods have been criticized for often failing to capture the model’s true reasoning~\citep{rudin2019stop,laugel2019dangers}, consequently offering limited guarantees for auditability, consistency or regulatory compliance. More recent efforts in neural NLP have attempted to incorporate explanations directly into the model’s output, for instance through self-explanation~\citep{nye2021show} or agentic feedback loops~\citep{madaan2023self}. However, these approaches rely on free-form natural language generation, which—although interpretable to humans—remains unverifiable and unstructured. As a result, they fall short of providing a consistent and auditable decision procedure.

\paragraph{Workflow and prompt optimization.}  
% Research on building trustworthy LLM and VLM-based systems has largely focused on workflow design and prompt refinement.
% Approaches such as AutoPrompt~\citep{shin2020autoprompt}, TextGrad~\citep{yuksekgonul2024textgrad}, and DSPy~\citep{khattab2023dspy} automatically optimize the prompt instructions to better align the LLM behavior with downstream tasks.
Research on building trustworthy LLM and VLM-based systems has largely focused on workflow design and prompt refinement, with approaches such as AutoPrompt~\citep{shin2020autoprompt}, TextGrad~\citep{yuksekgonul2024textgrad} and DSPy~\citep{khattab2023dspy} automatically optimizing prompt instructions to align LLM behavior with downstream tasks. These methods show that LLMs can be guided toward more accurate or systematic outputs without retraining, but do not provide explanation patterns.
Techniques such as Chain-of-Thought prompting~\citep{wei2022chain}, ReAct~\citep{yao2023react}, and Reflexion~\citep{shinn2023reflexion} structure reasoning through intermediate steps or self-feedback, providing both performance increases and transparent decision patterns. However, their reliance on free-form text generation makes them still vulnerable to hallucinations, questioning their efficacy in auditing and verification contexts.

We propose to address these challenges by enforcing a hierarchical tree-structured decision process over unstructured inputs. Defining each node as a natural-language question optimized through LLM feedback ensures both semantic adaptability and transparent reasoning paths. 
Our proposed framework, called Agentic Classification Tree,
thus aims to combine the interpretability of decision trees with the semantic reasoning of LLMs, providing an accountable alternative to existing black-box prompt optimization methods.

\section{Methodology}

To build a structured agentic workflow in the form of a classification tree, we take inspiration from traditional decision tree algorithms
such as CART~\citep{DBLP:books/wa/BreimanFOS84} and C4.5~\citep{quinlan1993c4}.
Table~\ref{tab:comparison-cart-act} provides a side-by-side
comparison between them %traditional decision trees
and our proposed ACT method.

\subsection{Traditional Decision Trees: CART and C4.5}

\begin{table}[tb]
  \caption{Comparison between traditional decision trees and ACT.} %(CART, C4.5) and our Agentic Classification Tree (ACT).}
  \label{tab:comparison-cart-act}
  \vskip 0.1in
  \centering
  \small
  \setlength{\tabcolsep}{4pt}
  \renewcommand{\arraystretch}{1.15}
  \begin{tabularx}{\columnwidth}{@{}>{\raggedright\arraybackslash}p{0.24\columnwidth}
                              >{\raggedright\arraybackslash}X
                              >{\raggedright\arraybackslash}X@{}}
    \toprule
      & \textbf{CART / C4.5} & \textbf{ACT (ours)} \\
    \midrule
    Data type
      & Tabular features
      & Unstructured inputs (text, images, etc.) \\
    \midrule
    Node definition
      & Numerical: binary threshold, Categorical: binary subset or multiway split
        %Cat.: CART binary subset; C4.5 multiway split
      & Binary natural- language question  queried by an LLM over the inputs. \\ % Binary natural- language question posed to an LLM. \\
    \midrule
    Best split search
      & Greedy search over candidate feature splits.
      & Iterative question refinement via \textsc{TextGrad}. \\ %(TextGrad) with LLM feedback. \\
    \midrule
    Split criterion
      & Gini (CART) or Information Gain Ratio (C4.5).
      & Gini impurity combined with semantic signals. \\ %(e.g., Gini/IG) augmented with semantic purity signals. \\
    \midrule
    Inference
      & Each input follows the tree based on feature values. % Route using observed feature values.
      & Each input is queried by the LLM at each node and routed according to its (yes/no) answers. \\ % Query the LLM at each node; route by its answer (yes/no). \\
    \bottomrule
  \end{tabularx}
  \vskip -0.1in
\end{table}

Classical decision-tree algorithms %such as CART~\citep{DBLP:books/wa/BreimanFOS84} and C4.5~\citep{quinlan1993c4} 
recursively partition a dataset based on feature values. 
At each node, a split is defined by either a threshold on a numerical feature or a subset of categorical values. 
The quality of each split is measured by an impurity criterion—typically Gini impurity (CART) or Information Gain ratio (C4.5)—and the algorithm greedily selects the split that yields the greatest impurity reduction, continuing recursively until a stopping criterion is met. %(e.g., maximum depth or node purity).
These methods are efficient and interpretable ~\citep{molnar2020interpretable} for structured, tabular data, but cannot operate directly on unstructured inputs such as text or images.
This limitation motivates our proposed ACT, which replaces feature-based splits with optimized natural language queries. 

\subsection{Proposed Approach: Agentic Classification Tree}
\label{sec:act_description}

%The Agentic Classification Tree (ACT)
% ACT is designed to extend classical decision tree methods to unstructured data. 
% Like CART and C4.5, ACT recursively partitions the input space, but instead of relying on numeric or categorical features, it leverages LLMs to define natural-language queries that guide data splits.

ACT is designed to extend classical decision tree methods to unstructured data by replacing numeric or categorical feature splits with LLM-generated natural-language questions that recursively partition the input space.

\paragraph{Node definition.}

Each node \( n \) in ACT is defined by a prompt \( p \) to semantically partition data via an LLM. Given an instance \( x_i \), the LLM is queried with \( p \) and responds with a binary answer (\texttt{yes} or \texttt{no}), determining the branch assignment. This process defines the following split:
\[
\begin{gathered}
\mathcal{D}_L^n = \{ (x_i, y_i) \in \mathcal{D}^n \mid f_{\text{split}}(p, x_i) = \texttt{yes} \}, \\
\mathcal{D}_R^n = \mathcal{D}^n \setminus \mathcal{D}_L^n.
\end{gathered}
\]

where \( f_{\text{split}}(p, x_i) \) denotes the LLM’s response to prompt \( p \) on input \( x_i \). The resulting subsets \( \mathcal{D}_L^n \) and \( \mathcal{D}_R^n \) thus reflect the semantic partition induced by the model at node \( n \). The splitting process is initialized with a neutral, generic question and in our experiments we use:

\begin{tcolorbox}[
  colback=yellow!10,
  colframe=yellow!35!black,
  boxrule=0.4pt,
  arc=2pt,
  left=4pt,right=4pt,top=2pt,bottom=2pt,
  boxsep=1pt,
  sharp corners=south,
]
\centering\small\textbf{Based on the provided example, does it belong to the positive class? (yes/no)}
\end{tcolorbox}

%This provides an unbiased starting point.
% Since \( f_{\text{split}} \) is neither trained on the task nor informed of the target labels, the initial partition is not expected to be meaningful. This is by design: withholding task-specific information ensures that decision prompts are derived from the training data rather than from prior knowledge. In contrast to approaches such as TextGrad~\citep{yuksekgonul2024textgrad}, which typically begin with a task-specific question and iteratively refine it, ACT starts from a neutral prompt and discovers meaningful questions from scratch through data-driven refinement.

Since \( f_{\text{split}} \) is neither trained on the task nor informed of the target labels, the initial partition is not expected to be meaningful. This is by design: withholding task-specific information ensures that decision prompts are derived from the training data rather than from prior knowledge. Unlike TextGrad~\citep{yuksekgonul2024textgrad}, which begins with a task-specific question and iteratively refines it, ACT starts from a neutral prompt and discovers meaningful questions through data-driven refinement.

\paragraph{Best split criterion.} 
ACT optimizes each decision node by iteratively refining a natural-language question \(p^{(k)}\) to achieve an effective semantic partition of the data. To guide this process, it combines quantitative evaluation (e.g., Gini impurity) with qualitative feedback from the LLM, yielding both statistical and semantic insights for prompt refinement. The procedure consists of two components:

\textbf{(1) Quantitative Impurity Evaluation.}  
As in classical decision tree algorithms, the quality of a split 
induced by the current prompt \(p^{(k)}\) is evaluated using standard impurity criteria $\delta(p^{(k)})$, e.g. weighted Gini impurity, %~\citep{DBLP:books/wa/BreimanFOS84}). % or information gain ratio~\citep{quinlan1993c4}).  We denote this score by \(\delta(p^{(k)})\), which serves as an objective function 
quantifying how well the semantic split separates the class labels.

\textbf{(2) Semantic Purity Analysis via LLM.} While the quantitative evaluation (e.g., Gini impurity) measures class-label heterogeneity within each partition, it does not reveal the \emph{semantic} factors underlying the impurity. 
To make these sources explicit, we analyze each child node independently by contrasting its class-conditional subsets (correct vs. misclassified examples). 
For this purpose, we align the predicted answer (\texttt{yes}/\texttt{no}) with the ground-truth label (\(y_i \in \{0,1\}\)) when defining correct and incorrect routing. 
Although this introduces an artificial correspondence between branch outcomes and class labels, it proved to be more effective in practice: the LLM more readily exploits feedback framed in terms of correct versus incorrect predictions than feedback based on contrasting the two branches directly (e.g., retaining the majority class and pruning the minority one). 
This alignment does not prevent the emergence of negative queries (e.g., ``\textit{Is feature \(A\) absent?}''), which the LLM can in principle generate and refine, though such formulations are typically harder to elicit consistently. Formally, each subset \( g \in \{\mathcal{D}_L^n, \mathcal{D}_R^n\} \) is associated with the model’s predicted answer %\(\hat{y}_g=1\)
\(\hat{y}_g\)
to the current prompt. For convenience, we denote by \(\hat{y}_g \in \{0,1\}\) the numeric encoding of this answer (\(\hat{y}_g=1\) for ``yes'', \(\hat{y}_g=0\) for ``no''). We then further partition each subset into two groups:

% \[
% X_{\mathrm{correct}}^{x},\; X_{\mathrm{error}}^{x}
% = \{(x_i,y_i)\in g \mid y_i \mathrel{\substack{=\\\neq}} \hat y_g\}
% \]

\[
\begin{aligned}
X_{\text{correct}}^n &= \{(x_i, y_i) \in g \mid y_i = \hat{y}_g\}, \\
X_{\text{error}}^n   &= \{(x_i, y_i) \in g \mid y_i \neq \hat{y}_g\}.
\end{aligned}
\]

Specifically, \( X_{\text{correct}} \) consists of input–label pairs where the LLM's predicted label \( \hat{y}_g \) matches the true label \( y_i \), while \( X_{\text{error}} \) consists of misclassified pairs where \(  y_i \ne \hat{y}_g \).

Next, we prompt the LLM to analyze these two groups and identify key \textbf{semantic characteristics} that distinguish them—i.e., features whose presence or absence could explain the misclassification of some examples. Based on this contrast, the LLM (denoted \(f_{\text{purity}}\)) returns concise, actionable feedback \(s_g\) that is then used to refine the node’s splitting question in the next optimization step to reduce impurity. By performing this semantic analysis independently on subsets \(D_L\) and \(D_R\), we obtain targeted feedback \(s_{D_L}\) and \(s_{D_R}\) that are then used to guide subsequent question-refinement iterations.

\noindent
\colorbox{blue!10}{%
\parbox{\dimexpr\columnwidth-2\fboxsep\relax}{%
    {\bfseries LLM Task: Semantic Purity Feedback for the \texttt{"yes"} group ($D_L$)\par}\smallskip

Below are two groups of samples for which a model answered either \texttt{"yes"} or \texttt{"no"} in response to a natural language prompt to predict their class label.

Provide feedback on key \textbf{characteristics} that are present or absent in the group where the model's predicted label matched the true label,  
and in the group where the prediction was incorrect.

For the following examples, the model answered \texttt{"yes"}:

\begin{itemize}\setlength\itemsep{0.2em}\setlength\parskip{0pt}\setlength\parsep{0pt}
    \item \textbf{Well-classified examples} (true label = \texttt{"yes"}):\\
    \emph{[List of correctly classified inputs]}
    
    \item \textbf{Misclassified examples} (true label = \texttt{"no"}):\\
    \emph{[List of misclassified inputs]}
\end{itemize}

The feedback you provide must be clear and concise.  
Focus on the one or two most important \textbf{characteristics}.
}}

\paragraph{Best split search.} 
At each node, the objective is to refine the prompt \(p^{(k)}\) so that the resulting partition minimizes the weighted Gini impurity \(\delta(p^{(k)})\). As described previously, this quantitative criterion is complemented by semantic feedback \((s_{D_L}, s_{D_R})\), obtained from LLM-based analyses of misclassified versus correctly classified examples. To jointly evaluate statistical impurity and semantic error patterns,
we define a semantic–impurity objective:
\[
\mathcal{L}(p^{(k)}) = f_{\mathrm{loss}}\!\big(p^{(k)}, s_{D_L}, s_{D_R}, \delta(p^{(k)})\big),
\]
where \(f_{\mathrm{loss}}\) is instantiated as an LLM. Given the current question, impurity statistics and semantic feedback, the LLM returns a natural-language diagnosis of its limitations, highlighting semantic factors that contribute to impurity.

We realize the refinement process using the differentiable prompting framework \textsc{TextGrad}~\citep{yuksekgonul2024textgrad}. Each refinement iteration proceeds in two stages:
\begin{enumerate}
    \item \emph{Guidance stage:} \texttt{TextGrad.feedback} takes the prompt \(p^{(k)}\) and its evaluation \(\mathcal{L}(p^{(k)})\) as input, and generates a natural-language editing instruction \(\nabla_p \mathcal{L}(p^{(k)})\) specifying how to revise the prompt.
    \item \emph{Revision stage:} \texttt{TextGrad.step} applies this instruction to produce an updated prompt:
    \[
    p^{(k+1)} = \texttt{TextGrad.step}\!\left(p^{(k)}, \nabla_p \mathcal{L}(p^{(k)})\right).
    \]
\end{enumerate}

The optimization loop runs for up to \(K\) steps or until convergence. 
At each iteration, the current prompt \(p^{(k)}\) is used to partition the data, evaluate impurity and analyze semantic error patterns via the LLM, as described above.
These signals jointly inform the loss \(\mathcal{L}(p^{(k)})\), which is then used by \texttt{TextGrad} to generate a refined prompt \(p^{(k+1)}\).
Among all candidate prompts \( p^{(0)}, \dots, p^{(K)} \), we select the one with the lowest training impurity \( p^\ast = \arg\min_k\, \delta(p^{(k)}) \).
%Among all prompts generated during the optimization process, \( p^{(0)}, p^{(1)}, \dots, p^{(K)} \), we select the one with the lowest impurity on the training set, i.e., \( p^\ast = \arg\min_k\, \delta(p^{(k)}) \).

\paragraph{Tree construction.}
As summarized in Algorithm~\ref{alg:prompt-cart}, ACT builds the decision tree recursively in a top-down manner. At each node, the best prompt is optimized via the procedure described above. If the resulting subset is pure or meets a predefined stopping criterion (e.g., minimum node size, %— the number of training examples in the node,
Gini impurity below a threshold, or maximum depth), a leaf node is created. As in classical decision trees, the leaf is assigned the majority class label among the examples it contains. %The complete ACT procedure is summarized in Algorithm~\ref{alg:prompt-cart}.

\paragraph{Tree interpretability.}
Decision trees, which rely on transparent structures and parameters to make decisions, are widely considered to be interpretable~\cite{molnar2020interpretable}. Owing to its shallow decision-tree structure and natural-language node questions, ACT exhibits the same desirable properties. In practice, interpretability depends on both the tree depth (i.e., the number of questions along a decision path) and the clarity of the questions. Accordingly, we restrict the tree depth to 2--4 to keep decision paths concise and explicitly instruct \textsc{TextGrad} during training to generate simple questions and limiting the number of logical clauses (\texttt{and}/\texttt{or}) to at most $L = 2$, as detailed in Appendix~\ref{appendix:prompt_constraints}.

% In practice interpretability depends on both the tree depth (i.e., the number of questions along a decision path) and the clarity of the questions. For this reason, we limit the tree depth to 3 -- 4 to keep decision paths concise and we explicitly instruct \texttt{TextGrad} during training to constrain generated questions to at most $L = 2$ logical clauses (AND/OR), as detailed in Appendix~\ref{appendix:prompt_constraints}.

\begin{algorithm}[tb]
\caption{ACT: Agentic Classification Tree}
\label{alg:prompt-cart}
\small
\begin{algorithmic} \\
{\bfseries Input:} Dataset $D$ \\
{\bfseries Output:} Decision tree $T$

% \Function{ACT}{$D$}
%   \State $T \gets$ \Call{GrowPromptTree}{$D$}
%   \State \Return $T$
% \EndFunction

\Function{GrowPromptTree}{$D$}
  \State Initialize prompt $p^{(0)}$ \Comment{default initialization}
  \If{$D$ is pure \textbf{or} stopping criterion is met}
    \State \Return Leaf node with majority class label
  \EndIf

  \For{$k = 0$ to $K-1$}
    \State $D_L \gets \{(x_i,y_i)\in D \mid f_{\text{split}}(p^{(k)},x_i)=\texttt{yes}\}$
    \State $D_R \gets D \setminus D_L$

    \State $\delta(p^{(k)}) \gets \frac{|D_L|}{|D|}\,\mathrm{Gini}(D_L)\;+\;\frac{|D_R|}{|D|}\,\mathrm{Gini}(D_R)$

    \ForAll{$(g,\hat{y}_g)\in\{(D_L,\texttt{yes}),(D_R,\texttt{no})\}$}
      \State $X_{\text{correct}} \gets \{(x_i,y_i)\in g \mid y_i=\hat{y}_g\}$
      \State $X_{\text{error}} \gets \{(x_i,y_i)\in g \mid y_i\neq \hat{y}_g\}$
      \State $s_g \gets f_{\text{purity}}(p^{(k)},X_{\text{correct}},X_{\text{error}})$
    \EndFor

    \State $\mathcal{L}(p^{(k)}) \gets f_{\text{loss}}(p^{(k)}, s_{D_L}, s_{D_R}, \delta(p^{(k)}))$
    \State $\nabla_p \mathcal{L}(p^{(k)}) \gets$ \Call{TextGrad.feedback}{$p^{(k)}, \mathcal{L}(p^{(k)})$}
    \State $p^{(k+1)} \gets$ \Call{TextGrad.step}{$p^{(k)}, \nabla_p \mathcal{L}(p^{(k)})$}
  \EndFor

  \State $p^\ast \gets \arg\min_{p^{(k)}} \delta(p^{(k)})$ \Comment{lowest impurity}
  \State Split $D$ using $p^\ast$ into $D_L$ and $D_R$
  \State node $\gets$ Create prompt node with final question $p^\ast$
  \State node.left $\gets$ \Call{GrowPromptTree}{$D_L$}
  \State node.right $\gets$ \Call{GrowPromptTree}{$D_R$}
  \State \Return node
\EndFunction

\Function{Gini}{$D$}
  \State \Return $1 - \sum_k p_k^2$ \Comment{$p_k$ = class proportion in $D$}
\EndFunction
\end{algorithmic}
\end{algorithm}

\section{Experiments}

\begin{table*}[!tb]
  \centering
  \caption{Comparison of classification methods across five datasets and four LLMs. Results show accuracy and F1-score on the test set (in \%) for six baselines, as well as the proposed ACT with varying hyperparameters ($d$: tree depth, $k$: optimization steps per node).}
  \label{tab:results_acart}

  \small
  \setlength{\tabcolsep}{4pt} % tighter columns (default is 6pt)
  \setlength{\arrayrulewidth}{0.6pt}
  
  \renewcommand{\arraystretch}{1.05} % slightly tighter rows without looking cramped

  \begin{tabular}{l rr|rr|rr|rr|rr}
    \toprule
    \multirow{2}{*}{\textbf{Method}}
      & \multicolumn{2}{c}{\textbf{Gemma3 4b}}
      & \multicolumn{2}{c}{\textbf{GPT-4.1 Nano}}
      & \multicolumn{2}{c}{\textbf{GPT-4.1 Mini}}
      & \multicolumn{2}{c}{\textbf{Qwen3 4b}}
      & \multicolumn{2}{c}{\textbf{Avg}} \\
    \cmidrule(lr){2-3} \cmidrule(lr){4-5} \cmidrule(lr){6-7} \cmidrule(lr){8-9} \cmidrule(lr){10-11}
      & \textbf{Acc} & \textbf{F1}
      & \textbf{Acc} & \textbf{F1}
      & \textbf{Acc} & \textbf{F1}
      & \textbf{Acc} & \textbf{F1}
      & \textbf{Acc} & \textbf{F1} \\
    \midrule

    %\multicolumn{11}{l}{\textbf{DIAGNO}} \\
    \textbf{DIAGNO} & & & & & & & & & & \\
    CoT (0-shot) & 61.5 & 69.5 & 63.3 & 54.7 & 61.2 & 51.4 & 63.2 & 41.7 & 62.3 & 54.3 \\
    DSPy (BFSR, 8 demos) & 64.0 & 68.2 & 68.7 & 61.2 & 64.5 & 59.4 & 63.3 & 54.9 & 65.1 & 60.9 \\
    \textsc{TextGrad} & 63.3 & 56.7 & 64.5 & 53.0 & 65.8 & 55.5 & 64.0 & 60.0 & 64.4 & 56.3 \\
    TF-IDF + CART ($d{=}3$) & $--$ & $--$ & $--$ & $--$ & $--$ & $--$ & $--$ & $--$ & $78.8$ & $78.9$ \\
    Con. tag. & 65.5 & 63.9 & 59.3 & 58.1 & 75.3 & 78.6 & 58.8 & 64.7 & 64.7 & 66.3 \\
    Rule Fit & $--$ & $--$ & $--$ & $--$ & $--$ & $--$ & $--$ & $--$ & $80.2$ & $79.3$ \\
    \textsc{ACT} ($d{=}3$, $k{=}10$) & 65.3 & 66.5 & 66.8 & 67.0 & 77.3 & 75.2 & 64.8 & 61.3 & 68.6 & 67.5 \\
    \textsc{ACT} ($d{=}4$, $k{=}20$) & \textbf{68.3} & \textbf{70.8} & \textbf{70.3} & \textbf{70.8} & \textbf{82.3} & \textbf{81.8} & \textbf{70.3} & \textbf{73.5} & 72.8 & 74.2 \\
    \midrule

    %\multicolumn{11}{l}{\textbf{SPAM}} \\
    \textbf{SPAM} & & & & & & & & & & \\
    CoT (0-shot) & 73.3 & 78.6 & 95.5 & 95.3 & 95.7 & 95.5 & 93.2 & 93.6 & 89.4 & 90.8 \\
    DSPy (BFSR, 8 demos) & 95.0 & 95.2 & 98.3 & 98.3 & 98.2 & 98.2 & 97.2 & 97.2 & 97.2 & 97.2 \\
    \textsc{TextGrad} & 95.0 & 94.7 & 97.3 & 97.3 & 96.7 & 96.6 & 96.0 & 96.0 & 96.3 & 96.2 \\
    TF-IDF + CART ($d{=}5$) & $--$ & $--$ & $--$ & $--$ & $--$ & $--$ & $--$ & $--$ & $92.7$ & $93.1$ \\
    Con. tag. & 83.2 & 85.5 & 64.2 & 45.8 & 82.3 & 85.0 & 62.3 & 39.6 & 73.0 & 64.0 \\
    Rule Fit & $--$ & $--$ & $--$ & $--$ & $--$ & $--$ & $--$ & $--$ & $97.0$ & $96.9$ \\
    \textsc{ACT} ($d{=}2$, $k{=}5$) & 95.8 & 95.8 & 98.7 & 98.6 & 98.2 & 98.2 & 96.7 & 96.6 & 97.4 & 97.3 \\
    \textsc{ACT} ($d{=}3$, $k{=}10$) & \textbf{98.5} & \textbf{98.5} & \textbf{99.2} & \textbf{99.2} & \textbf{99.5} & \textbf{99.5} & \textbf{98.8} & \textbf{98.8} & 99.0 & 99.0 \\
    \midrule

    %\multicolumn{11}{l}{\textbf{JAILBREAK}} \\
    \textbf{JAILBREAK} & & & & & & & & & & \\
    CoT (0-shot) & 77.9 & 82.2 & 85.5 & 83.5 & 93.6 & 93.3 & 81.1 & 77.7 & 84.5 & 84.2 \\
    DSPy (BFSR, 8 demos) & 89.6 & 89.3 & 91.6 & \textbf{91.6} & 94.8 & 94.8 & 85.5 & 83.9 & 90.4 & 89.9 \\
    \textsc{TextGrad} & \textbf{91.6} & \textbf{91.3} & 90.4 & 90.3 & 94.8 & 95.1 & 88.8 & 90.0 & 91.4 & 91.7 \\
    TF-IDF + CART ($d{=}4$) & $--$ & $--$ & $--$ & $--$ & $--$ & $--$ & $--$ & $--$ & $92.8$ & $92.4$\\
    Con. tag. & 82.7 & 82.2 & 58.6 & 31.8 & 82.7 & 79.8 & 76.3 & 70.1 & 75.1 & 66.0 \\
    Rule Fit & $--$ & $--$ & $--$ & $--$ & $--$ & $--$ & $--$ & $--$ & $96.8$ & $96.8$ \\
    \textsc{ACT} ($d{=}3$, $k{=}10$) & 82.3 & 82.7 & 85.5 & 85.1 & 95.2 & 95.5 & 82.7 & 82.6 & 86.4 & 86.5 \\
    \textsc{ACT} ($d{=}4$, $k{=}20$) & 90.8 & 90.0 & \textbf{92.0} & 91.5 & \textbf{98.8} & \textbf{98.8} &\textbf{92.0} & \textbf{92.7} & 93.4 & 93.3 \\
    \midrule

    %\multicolumn{11}{l}{\textbf{BANKCHURN}} \\
    \textbf{BANKCHURN} & & & & & & & & & & \\
    CoT (0-shot) & 47.8 & 48.1 & 48.5 & 25.9 & 52.8 & 60.5 & 50.2 & 37.3 & 49.8 & 43.0 \\
    DSPy (BFSR, 8 demos) & 52.8 & 55.6 & 52.3 & 42.8 & 57.2 & 58.5 & 54.0 & 52.2 & 54.1 & 52.3 \\
    \textsc{TextGrad} & 52.0 & 60.0 & 53.3 & 60.3 & 55.3 & 61.0 & 52.8 & 61.4 & 53.4 & 60.7 \\
    TF-IDF + CART ($d{=}5$) & $--$ & $--$ & $--$ & $--$ & $--$ & $--$ & $--$ & $--$ & $63.2$ & $66.8$\\
    Con. tag. & 57.2 & 60.2 & 52.0 & 59.8 & 59.7 & 60.8 & 54.3 & 61.6 & 55.8 & 60.6 \\
    Rule Fit & $--$ & $--$ & $--$ & $--$ & $--$ & $--$ & $--$ & $--$ & $63.0$ & $59.8$ \\
    \textsc{ACT} ($d{=}3$, $k{=}10$) & 56.3 & 42.3 & 58.8 & 67.4 & 60.5 & 64.6 & 57.5 & \textbf{62.7} & 58.3 & 59.3 \\
    \textsc{ACT} ($d{=}4$, $k{=}20$) & \textbf{65.2} & \textbf{70.7} & \textbf{65.0} & \textbf{69.1} & \textbf{68.7} & \textbf{68.6} & \textbf{62.0} & 60.0 & 65.2 & 67.1 \\
    \midrule

    %\multicolumn{11}{l}{\textbf{IMDB}} \\
    \textbf{IMDB} & & & & & & & & & & \\
    CoT (0-shot) & 92.3 & 91.9 & 94.2 & 94.1 & 95.7 & 95.6 & 94.2 & 94.0 & 94.1 & 93.9 \\
    DSPy (BFSR, 8 demos) & \textbf{93.7} & \textbf{93.6} & \textbf{95.5} & \textbf{95.5} & \textbf{96.8} & \textbf{96.8} & 94.8 & 94.8 & 95.2 & 95.2 \\
    \textsc{TextGrad} & 93.2 & 92.9 & 95.2 & 95.1 & 95.8 & 95.8 & 94.7 & 94.5 & 94.7 & 94.6 \\
    TF-IDF + CART ($d{=}3$) & $--$ & $--$ & $--$ & $--$ & $--$ & $--$ & $--$ & $--$ & $66.8$ & $72.8$\\
    Con. tag. & 74.8 & 76.9 & 56.2 & 31.7 & 50.0 & 66.7 & 51.3 & 63.8 & 58.1 & 59.8 \\
    Rule Fit & $--$ & $--$ & $--$ & $--$ & $--$ & $--$ & $--$ & $--$ & $75.2$ & $73.7$ \\
    \textsc{ACT} ($d{=}3$, $k{=}10$) & 93.2 & 93.2 & 92.8 & 92.8 & 95.7 & 95.7 & 93.8 & 93.9 & 93.9 & 93.9 \\
    \textsc{ACT} ($d{=}3$, $k{=}20$) & \textbf{93.7} & \textbf{93.6} & 94.7 & 94.7 & 96.3 & 96.2 & \textbf{95.5} & \textbf{95.5} & 95.1 & 95.0 \\
    \bottomrule
  \end{tabular}
\end{table*}

We empirically evaluate ACT %the proposed Agentic Classification Tree (ACT)
against state-of-the-art baseline methods on multiple binary text classification datasets.

\subsection{Setup: Datasets, Models \& Baselines}

\paragraph{Datasets.}
We evaluate ACT on five binary text-classification tasks: %spanning medical diagnosis, content moderation, sentiment analysis, and structured customer-attribute prediction. 
The \textbf{DIAGNO} dataset contains short clinical descriptions labeled as tuberculosis (TB) or not TB, and we use balanced train–test splits for controlled evaluation. 
\textbf{SPAM} consists of email messages annotated as spam or ham, also evaluated with balanced splits. 
\textbf{JAILBREAK} contains user prompts labeled according to whether they attempt to circumvent safety policies, and is used in its original moderately imbalanced form.
\textbf{BANKCHURN} is a customer-churn dataset in which tabular attributes are serialized into short textual profiles following the protocol used by~\citep{arzaghi2025intrinsic}. This task emphasizes dataset-specific attribute relationships rather than broad world knowledge. 
% \textbf{IMDB} is a standard sentiment dataset of movie reviews, from which we use a balanced subset to maintain consistency across tasks.
\textbf{IMDB} is a standard sentiment dataset of movie reviews; we use a balanced subset for consistency.

%\paragraph{Models.}
\noindent\textbf{Models.} We evaluate our method on four language models: Gemma-4B~\cite{team2025gemma}, GPT-Nano-4.1~\cite{achiam2023gpt}, GPT-Mini-4.1~\cite{achiam2023gpt}, and Qwen3-4B~\cite{yang2025qwen3}. These models were selected to represent diverse architectural approaches while prioritizing computationally efficient smaller models. To assess the effect of model scale, %we run all main experiments on the Nano and Mini variants of GPT-4.1 and
we report additional scaling results with full GPT-4.1~\cite{achiam2023gpt}.

%\paragraph{Baselines.}
\noindent\textbf{Baselines.} We evaluate ACT against six baselines: \textbf{Chain-of-Thought (CoT, zero-shot)}~\citep{wei2022chain}. %uses standard zero-shot prompting with step-by-step reasoning, without access to labeled examples. %or demonstrations.
\textbf{DSPy (BFSR, 8 demos)}~\citep{khattab2023dspy} following the setup of~\citep{yuksekgonul2024textgrad}: Bootstrapped Few-Shot Random Search (BFSR) generates reasoning traces and answers for all training samples, filters for correct answers, constructs candidate prompts with up to 8 correct demonstrations (input, reasoning trace and output), evaluates the candidate prompts on the whole training set and selects the best-performing one. \textbf{TextGrad}~\citep{yuksekgonul2024textgrad} refines a task-specific prompt through textual feedback, optimizing for accuracy given the task description and class labels. We use a batch size of 50 samples and run 12 epochs with 3 optimization steps each, yielding 36 refinement iterations where each candidate is evaluated on a validation set and the best is retained. \textbf{TF-IDF + CART} trains a CART classifier (depths 3–5) on TF-IDF text representations~\citep{salton1975vector,DBLP:books/wa/BreimanFOS84}, exhaustively searching for optimal splits without relying on a language model. This offers structural transparency but limited semantic interpretability. The best-performing depth per dataset is reported in Table~\ref{tab:results_acart}, with full results in Appendix~\ref{appendix:classical_baselines}.
\textbf{LLM Concept Bottleneck + CART (Con.\ tag.)} is a concept-bottleneck baseline where an LLM extracts salient domain concepts and encodes each text as a binary concept vector, on which a shallow CART is then trained. \textbf{RuleFit (TF-IDF rules)}~\citep{friedman2008predictive} is a rule-ensemble method that learns sparse, human-readable if--then rules over TF-IDF features.
\looseness=-1

\paragraph{Configurations.} The three prompt-based baseline methods (CoT, DSPy, TextGrad) are initialized with task-specific prompts that explicitly state the classification objective and define the target classes (e.g., for DIAGNO, whether a case indicates Tuberculosis). In contrast, ACT begins from a generic, task-agnostic query (``Based on the provided example, does it belong to the positive class?''), without access to the task description or class names. %, or any domain-specific information.
Consequently, ACT must uncover the task through successive node's question-refinement iterations, making the comparison conservative with respect to considered baselines. We vary two hyperparameters of the proposed ACT algorithm: the tree depth \(d\) and the number of prompt refinement steps per node \(k\). %exploring different configurations in our experiments.
To control input length and computational cost, \(f_{\text{purity}}\) uses at most \(m\) randomly selected well-classified and \(m\) randomly selected misclassified instances per group (with \(m=50\) in our experiments). %To ensure a fair comparison, all methods are evaluated using identical underlying language models.

\subsection{Experimental Results and Discussion}

\paragraph{Test set accuracy} As shown in Table~\ref{tab:results_acart}, \textsc{ACT} with appropriately selected hyperparameters consistently matches or surpasses the four LLM-based baselines (CoT, \textsc{TextGrad}, Con. Tag. and DSPy) across datasets and LLMs. Averaged over all datasets and LLMs, the best-performing ACT configuration %($d=4$, $k=20$)
improves test accuracy by $4.7$ percentage points over DSPy and by $5.1$ points over \textsc{TextGrad}, while providing %transparent and
interpretable decision paths. When evaluated by F1-score, ACT outperforms all four LLM-based baselines in 15 of the 20 dataset-model configurations.

Compared to TF-IDF + CART, ACT consistently achieves similar or higher accuracy on all tasks except for DIAGNO, where only the GPT-4.1-mini variant surpasses this baseline. However, TF-IDF+CART remains fundamentally limited by its lack of semantic interpretability. The resulting decision trees (cf. Figures~\ref{fig:tree-tfidf} and~\ref{fig:tree-tfidf-spam}) are typically difficult to interpret compared to the natural-language questions learned by ACT (cf. Figures~\ref{fig:full_combined_tb},~\ref{fig:spam_tree_qwen3}, and~\ref{fig:jailbreak_tree_qwen3}). This stems from TF-IDF+CART selecting individual words or n-grams as decision criteria without regard to context or semantic meaning, causing the model to struggle with negation, irony, rephrasing and adversarially crafted text—features particularly prevalent in SPAM, JAILBREAK and IMDB.
% Lastly, ACT with GPT-4.1-mini surpasses RuleFit on all tasks, but RuleFit outperforms ACT on average for DIAGNO and JAILBREAK. A key advantage of ACT over TF-IDF+CART and RuleFit is its modularity: since ACT can be built on any LLM accessible through an API, its performance scales with advances in language model development and practitioners can select models suited to their specific accuracy or cost requirements.
Compared to RuleFit, ACT with GPT-4.1-mini consistently achieves higher accuracy on all tasks, but RuleFit outperforms ACT on average for DIAGNO and JAILBREAK. Like TF-IDF + CART, RuleFit offers limited interpretability because its rules operate on TF-IDF embeddings (e.g., nasal $\le 0.238$), whereas ACT provides human-readable yes/no questions. Additional experiments (Appendix~\ref{appendix:classical_baselines}) show that ACT remains competitive with strong non-interpretable baselines (XGBoost, BERT, RoBERTa) while providing interpretable and human-readable decision paths.

% Additionally, we compare ACT to strong non-interpretable baselines: XGBoost trained on TF-IDF representations~\citep{chen2016xgboost}, fine-tuned BERT~\citep{devlin2019bert} and RoBERTa~\citep{liu2019roberta} classifiers.As shown in Appendix~\ref{appendix:classical_baselines}, these models achieve strong accuracy, though ACT remains competitive while providing interpretable decision paths. %explicit, human-readable decision paths.

\begin{table}[b]
  \caption{\textbf{Interpretability Statistics} averaged over the 4 LLMs evaluated on DIAGNO. For reference, the generic seed question \textit{“Based on ... positive class? (yes/no)”} has a length of 87 characters.}
  \label{tab:interpret-diagno-main}
  \begin{center}
    \begin{small}
      \begin{sc}
        \begin{tabular}{lcc}
          \toprule
            ACT ($d$, $k$) & Avg. Q. Len & Avg. Path \\
            \midrule
            (3, 10) & 92.00 & 2.35 \\
            (3, 20) & 93.15 & 2.51 \\
            (4, 10) & 78.05 & 3.03 \\
            (4, 20) & 80.53 & 3.06 \\
            \bottomrule
        \end{tabular}
      \end{sc}
    \end{small}
  \end{center}
  \vskip -0.1in
\end{table}

\paragraph{Interpretability.}
To quantify ACT's interpretability, we report the average path length at inference and average node-question length in Table~\ref{tab:interpret-diagno-main} (full results in Appendix~\ref{appendix:interpretability}). Predictions require only about three nodes even for depth-4 trees and questions remain easily understandable ($\le$ 155 characters), indicating that soft constraints on question complexity, particularly the limit $L$ on logical clauses, are effective. These quantitative results, together with manual inspection of generated trees (Figure~\ref{fig:full_combined_tb}; Appendix~\ref{appendix:act_spam_jailbreak}), confirm that ACT's divide-and-conquer approach of decomposing complex classification tasks into simpler subquestions yields interpretable decision paths while improving accuracy. %compared to other prompt optimizaton methods.
Furthermore, Table~\ref{tab:compare-act-diagno} compares the symptoms for Tuberculosis identified by ACT with established medical sources ~\cite{who2024tuberculosis}, revealing strong alignment and demonstrating ACT's efficacy in high-stakes scenarios (cf.\ Appendix~\ref{sec:comparison-experts}).

\begin{table}[tb]
  \caption{Symptoms comparison between ACT and medical sources.}
  \label{tab:compare-act-diagno}
  \begin{center}
    \begin{small}
      \begin{sc}
        \begin{tabular}{@{}p{0.95\columnwidth}@{}}
          \toprule
          \textbf{In common (5):} Coughing up blood, Fever, Weight loss, Swollen lymph nodes, Fatigue \\
          \midrule
          \textbf{ACT only (2):} Shortness of breath, Severe pain \\
          \midrule
          \textbf{Medical sources only (4):} Chest pain, Headache, Increased sweating, Loss of appetite \\
          \bottomrule
        \end{tabular}
      \end{sc}
    \end{small}
  \end{center}
  \vskip -0.1in
\end{table}

% \begin{figure}[tb]
%   \vskip 0.2in
%   \begin{center}
%   \centerline{\includegraphics[width=\columnwidth]{images/diagno_ablation_k_icml_main.pdf}}

%   \caption{\textbf{Ablation study of ACT hyperparameter $k$} on the DIAGNO dataset using GPT-4.1 Nano.} %Solid lines denote training accuracy; dashed lines denote test accuracy.}
%   \label{fig:ablation_diagno}
%   \end{center}
%     \vskip -0.2in
% \end{figure}

\begin{figure}[tb]
  \vskip 0.2in
  \begin{center}
  \includegraphics[width=\columnwidth]{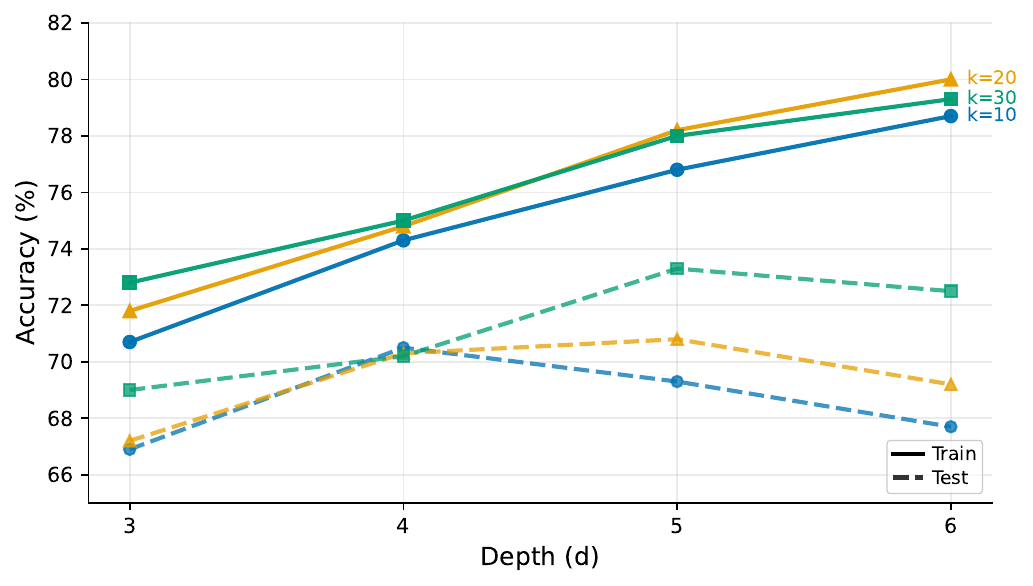}
  \vskip 0.1in
  \includegraphics[width=\columnwidth]{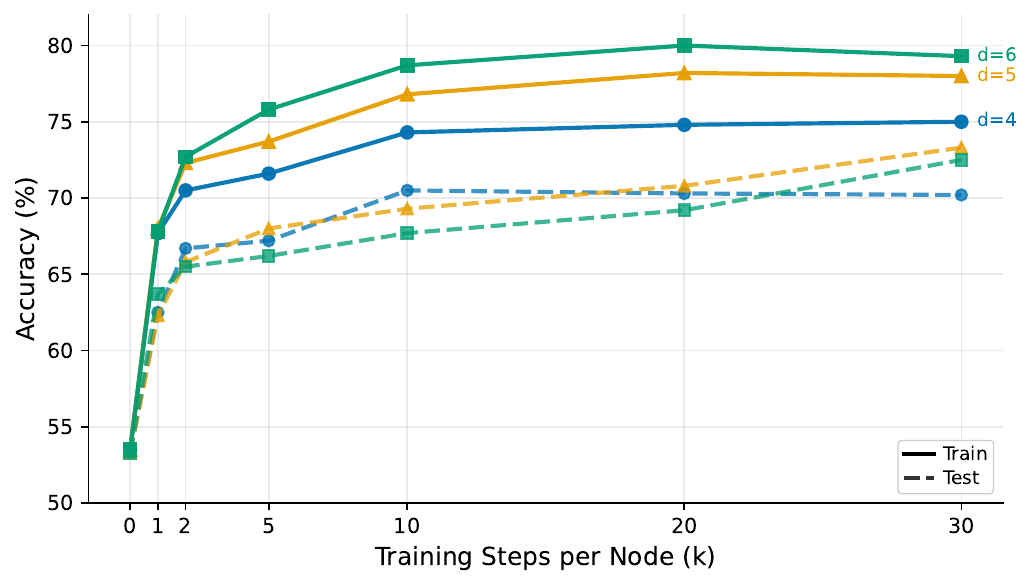}
  \caption{\textbf{Ablation study of ACT hyperparameters} on DIAGNO using GPT-4.1 Nano. Top: effect of tree depth $d$. Bottom: effect of optimization steps $k$.}
  \label{fig:ablation_diagno}
  \end{center}
  \vskip -0.1in
\end{figure}

% \paragraph{Ablation studies.} ACT has two key hyperparameters: the tree depth $d$ and the number of optimization steps per node $k$. As illustrated in Figure~\ref{fig:ablation_diagno}, both $d$ and $k$ significantly influence ACT performance. Across datasets and LLMs, test accuracy peaks at depths 4--5, with deeper trees exhibiting decreased test accuracy despite increased training accuracy at depth 6---an indication of overfitting. Conversely, increasing optimization steps consistently improves performance, though gains plateau after 10--20 steps, indicating that a modest number of steps suffices to identify the best splitting question.

% \paragraph{Ablation studies.} ACT has two key hyperparameters: tree depth $d$ and number of optimization steps per node $k$. As shown in Figure~\ref{fig:ablation_diagno}, both significantly influence ACT's performance. For DIAGNO test accuracy peaks at depths 4--5, with deeper trees exhibiting decreased test accuracy despite increased training accuracy at depth 6 - an indication of overfitting. Increasing $k$ consistently improves performance, though gains plateau after 10--20 steps, indicating that a modest number of steps suffices to identify the best splitting question.
% Additional results, including on the number $m$ of examples from $X_{\text{correct}}$ and $X_{\text{error}}$ provided to the LLM ($f_{\text{purity}}$) during semantic analysis are in Appendix~\ref{appendix:add_ablations}.

\paragraph{Ablation studies.} As shown in Figure~\ref{fig:ablation_diagno}, tree depth $d$ and optimization steps per node $k$ both significantly influence ACT's performance. As expected for tree-based methods, increasing depth beyond a certain point (e.g., from 5 to 6 for DIAGNO with GPT-4.1-Nano) leads to overfitting. Increasing $k$ consistently improves performance, though gains plateau after 10--20 steps. Additional results, including ablations on the number $m$ of examples provided to $f_{\text{purity}}$ during semantic analysis, are in Appendix~\ref{appendix:add_ablations}.

\begin{figure}[tb]
  \vskip 0.2in
  \begin{center}
  \centerline{\includegraphics[width=\columnwidth]{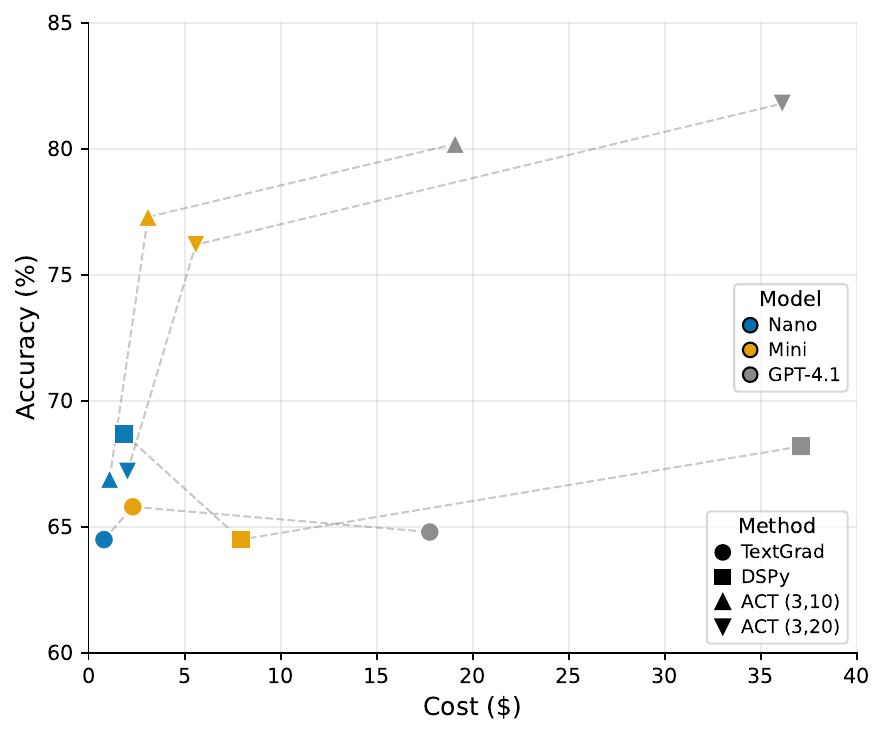}}
  % \caption{\textbf{Scaling analysis across the GPT-4.1 family.} The graph shows the test set accuracy in \% and training cost in US-\$ for ACT $(3, 10)$, ACT $(3, 20)$, DSPy and \textsc{TextGrad} for \textcolor{blue}{GPT-4.1-Nano}, \textcolor{orange}{GPT-4.1-Mini} and \textcolor{gray}{GPT-4.1} on the DIAGNO dataset. Cost denotes training expense using Azure-hosted GPT-4.1 models.}
  \caption{\textbf{Scaling analysis across the GPT-4.1 family.} Test accuracy (\%) versus training cost (USD) for \textsc{TextGrad}, DSPy, ACT $(d=3, k=10)$ and ACT $(d=3, k=20)$ on DIAGNO using Azure-hosted GPT-4.1 models.}
  \label{fig:ablation_diagno_cost}
  \end{center}
\vskip -0.2in
\end{figure}

\paragraph{Scaling analysis.} We evaluate how ACT performance scales with model capacity across the GPT-4.1 family (Nano, Mini and full GPT-4.1). As shown in Figure~\ref{fig:ablation_diagno_cost}, ACT's test accuracy consistently improves with stronger models, whereas DSPy and TextGrad do not exhibit consistent scaling behavior, indicating that ACT more effectively leverages the enhanced reasoning capabilities of larger models. This observed scaling trend also suggests a structural advantage of ACT over classical tree-based baselines such as TF-IDF + CART and RuleFit: because ACT can be instantiated with any API-accessible LLM, its performance benefits directly from advances in language model development.

% This observed scaling trend also illustrates a structural difference between ACT and classical tree-based baselines such as TF-IDF + CART and RuleFit: because ACT can be instantiated with any API-accessible LLM, its performance benefits directly from advances in language model development.

%As detailed in Appendix~\ref{app:comp_cost_analysis}, we compare wall-clock time and token usage for TextGrad, DSPy and ACT using Gemma-3-4B. ACT $(d=3,k=10)$ achieves similar or lower cost than the baselines, while ACT $(d=4,k=20)$ is more expensive but provides improved accuracy and interpretability.

\section{Conclusion}\label{sec:conclusion}
We introduced the Agentic Classification Tree (ACT), a novel framework that combines the interpretability of traditional decision trees with the semantic reasoning capabilities of LLMs. Unlike conventional decision trees that rely on rigid feature-based splits, ACT dynamically optimizes natural-language prompts at each node, using LLM responses to perform semantically meaningful binary splits. Our experiments on text-based binary classification tasks demonstrate that ACT achieves competitive accuracy while providing interpretable, language-based decision logic. This approach highlights the potential of combining classical machine learning structures with modern language model reasoning, opening new avenues for interpretable and effective decision-making in complex, text-rich domains. Future work includes extending ACT to multi-class classification and regression tasks, improving computational efficiency through targeted prompt optimization strategies, and exploring applications beyond text classification to other structured and unstructured data modalities.


% Authors are kindly asked to make their submissions as accessible as possible
% for everyone including people with disabilities and sensory or neurological
% differences. Tips of how to achieve this and what to pay attention to will be
% provided on the conference website \url{http://icml.cc/}.

% \section*{Software and Data}

% If a paper is accepted, we strongly encourage the publication of software and
% data with the camera-ready version of the paper whenever appropriate. This can
% be done by including a URL in the camera-ready copy. However, \textbf{do not}
% include URLs that reveal your institution or identity in your submission for
% review. Instead, provide an anonymous URL or upload the material as
% ``Supplementary Material'' into the OpenReview reviewing system. Note that
% reviewers are not required to look at this material when writing their review.

% \section*{Acknowledgements}

% \textbf{Do not} include acknowledgements in the initial version of the paper

% If a paper is accepted, the final camera-ready version can (and usually should)
% include acknowledgements.  Such acknowledgements should be placed at the end of
% the section, in an unnumbered section that does not count towards the paper
% page limit. Typically, this will include thanks to reviewers who gave useful
% comments, to colleagues who contributed to the ideas, and to funding agencies
% and corporate sponsors that provided financial support.

\section*{Impact Statement}

% Authors are \textbf{required} to include a statement of the potential broader
% impact of their work, including its ethical aspects and future societal
% consequences. This statement should be in an unnumbered section at the end of
% the paper (co-located with Acknowledgements -- the two may appear in either
% order, but both must be before References), and does not count toward the paper
% page limit. In many cases, where the ethical impacts and expected societal
% implications are those that are well established when advancing the field of
% Machine Learning, substantial discussion is not required, and a simple
% statement such as the following will suffice:

% ``This paper presents work whose goal is to advance the field of Machine
% Learning. There are many potential societal consequences of our work, none
% which we feel must be specifically highlighted here.''

% The above statement can be used verbatim in such cases, but we encourage
% authors to think about whether there is content which does warrant further
% discussion, as this statement will be apparent if the paper is later flagged
% for ethics review.

Algorithmic decision making has a broad range of ethical implications, for example, allocative harms or representational harms \cite{shelby2023sociotechnicalharmsalgorithmicsystems}. Bias in algorithmic decision making especially often harms minorities \cite{lum-to-predict-and-serve}, \cite{pmlr-v81-buolamwini18a}, \cite{obermeyer-racial-bias}. To reduce these harms and for other reasons, recent governance frameworks and regulations such as the EU AI Act\footnote{\url{https://artificialintelligenceact.eu/}}, the OECD AI Principles\footnote{\url{https://oecd.ai/en/ai-principles}} and the NIST AI Risk Management Framework\footnote{\url{nist.gov/itl/ai-risk-management-framework}} emphasize that AI-based decisions in high-stakes scenarios must be explainable and subject to human oversight. By providing explicit, interpretable decision paths that are accessible not only to technical experts but also to stakeholders, auditors and regulators, ACT offers an algorithmic decision-making framework that combines competitive accuracy with transparency and contestability. 

Nonetheless, two important limitations must be considered. First, while ACT reduces reasoning errors, it remains LLM-based and therefore may still exhibit hallucinations, inconsistencies or unreliability. %Second, as noted by \cite{oneil-wmp}, transparency alone does not guarantee effective contestability: meaningful recourse requires access to appropriate institutional or regulatory bodies.
Second, as noted by \cite{oneil-wmp}, transparency and contestability, even if guaranteed by the algorithm, are only effective in practice when embedded within appropriate institutional or regulatory frameworks that provide meaningful avenues for recourse. We therefore urge practitioners deploying ACT in real-world settings to ensure adequate human oversight and mechanisms that allow decisions to be contested in a timely and effective manner.

\bibliography{example_paper}
\bibliographystyle{icml2026}

%%%%%%%%%%%%%%%%%%%%%%%%%%%%%%%%%%%%%%%%%%%%%%%%%%%%%%%%%%%%%%%%%%%%%%%%%%%%%%%
%%%%%%%%%%%%%%%%%%%%%%%%%%%%%%%%%%%%%%%%%%%%%%%%%%%%%%%%%%%%%%%%%%%%%%%%%%%%%%%
% APPENDIX
%%%%%%%%%%%%%%%%%%%%%%%%%%%%%%%%%%%%%%%%%%%%%%%%%%%%%%%%%%%%%%%%%%%%%%%%%%%%%%%
%%%%%%%%%%%%%%%%%%%%%%%%%%%%%%%%%%%%%%%%%%%%%%%%%%%%%%%%%%%%%%%%%%%%%%%%%%%%%%%
\newpage
\appendix

\onecolumn

\section{Computational Cost Analysis}
\label{app:comp_cost_analysis}

\begin{figure}[htb]

\centering

    \centering
    \includegraphics[width=\textwidth]{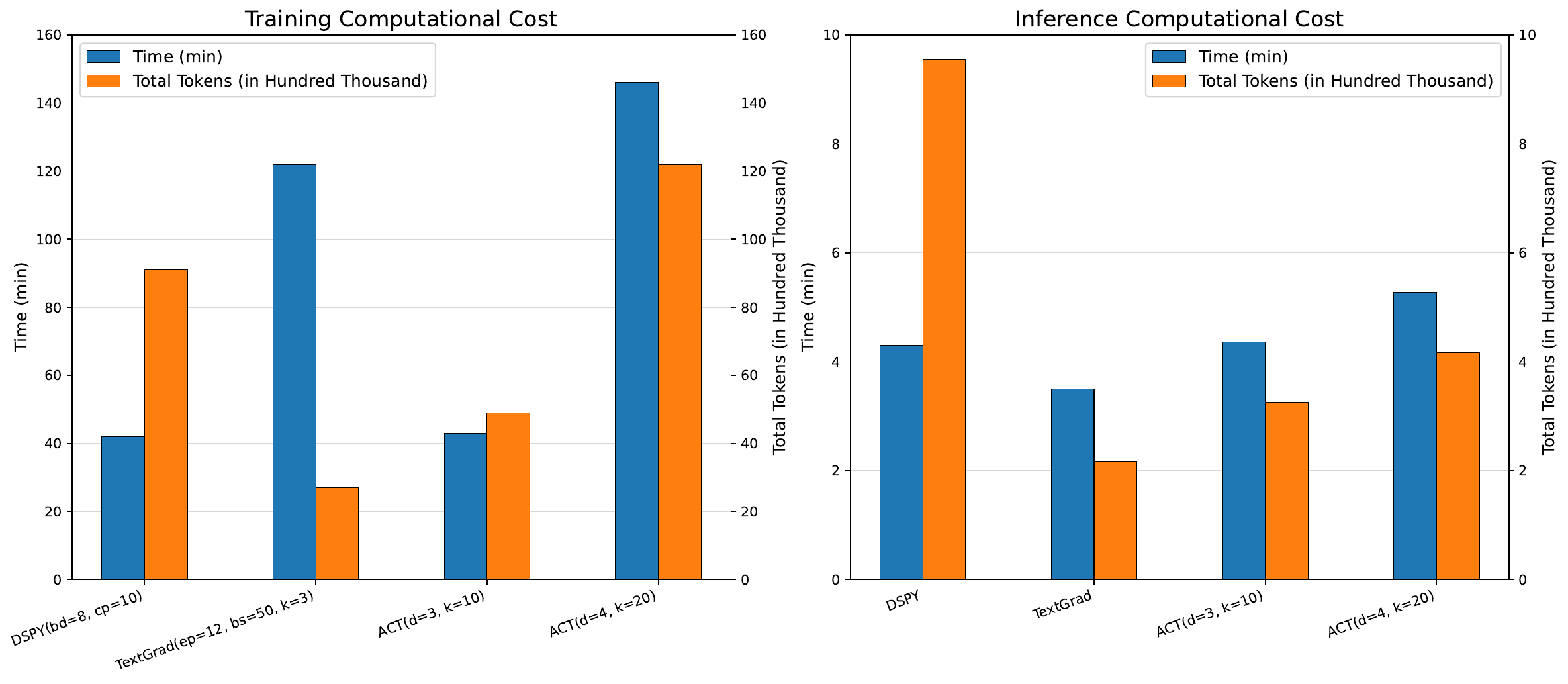}
    \caption{\textbf{Training and inference computational cost comparison on DIAGNO.} 
    Each plot reports the average wall-clock time and token usage over 5 runs for TextGrad, DSPy (BFSR, 8-shot), and ACT under both the cheapest (d = 3, k = 10) and most expensive (d = 4, k = 20) configurations, using Gemma-3-4B on a single H200 GPU with 8 threads per program. 
    Training cost reflects program construction dynamics: TextGrad processes fewer but very long prompts; DSPy evaluates 13 medium-sized candidate programs derived from few-shot traces; and ACT issues many short node-level prompts for Gini estimation supplemented with a smaller number of larger prompts for semantic feedback and node optimization. Inference cost shows that TextGrad is the most economical, while ACT remains within the same order of magnitude and provides higher accuracy with more interpretable outputs. DSPy exhibits similar wall-clock time to ACT but requires more tokens, largely due to the substantial size of the few-shot prompts.}

    \label{fig:comp_cost_diagno}
\end{figure}

\paragraph{Computational cost analysis.}
Figure~\ref{fig:comp_cost_diagno} compares the training and inference costs of DSPy, TextGrad and two ACT configurations on the DIAGNO task. We report wall-clock time and total token usage (input plus output). During training, all methods have access to the full training set and at inference time all methods are evaluated on the full test set. All experiments were conducted using Gemma-3-4B-IT on a single H200 GPU with 8 threads per program and results are averaged over five independent runs.

The observed differences reflect the underlying program-construction strategies of each method. TextGrad exhibits relatively high wall-clock time despite lower token usage, as it relies on fewer but substantially longer prompts. DSPy, by contrast, evaluates a larger number of medium-sized prompts derived from few-shot reasoning traces. ACT issues many short prompts to estimate Gini impurity at each node, complemented by a smaller number of longer prompts for semantic feedback and node-level optimization. In its cheapest configuration, ACT is competitive with both baselines in overall computational cost. Increasing tree depth raises training cost but yields higher accuracy while preserving interpretability. At inference time, TextGrad is the most economical, while ACT remains within the same order of magnitude and provides higher accuracy with interpretable decision paths. DSPy shows similar wall-clock time to ACT but incurs higher token usage due to the substantial size of its few-shot prompts.

\newpage

\section{Additional Implementation Details of the ACT Algorithm}

\subsection{Illustrative Example: Prompt Refinement via Semantic Feedback}
\label{sec:prompt_refinement_example}

We describe the iterative prompt refinement procedure used in ACT, using an illustrative case from the \texttt{DIAGNO} medical diagnosis task. At each decision node, the objective is to construct a binary natural-language question that separates the data into semantically meaningful groups, thereby reducing weighted Gini impurity. At each iteration \(k\), the following steps are performed:
\begin{enumerate}
    \item The current prompt \(p^{(k)}\) is evaluated over the training set. For each input \(x_i\), the LLM outputs a binary decision (\texttt{yes}/\texttt{no}).
    \item The data is partitioned into two subsets: those routed to the \texttt{yes} branch (denoted \textbf{$sd_L$}) and those routed to the \texttt{no} branch (denoted \textbf{$sd_G$}).
    \item For each subset, the correctly classified and misclassified examples are separated. These are then passed to the LLM, which is prompted to identify semantic features that distinguish the two groups.
    \item Based on this contrastive analysis, the LLM generates feedback indicating which aspects of the current prompt are ambiguous or insufficient. This feedback is then used to revise the question, yielding a new candidate \(p^{(k+1)}\).
\end{enumerate}

This iterative refinement continues until a stopping criterion is met (e.g., maximum number of updates). In practice, we observe that prompts evolve from generic task-agnostic formulations to more specific, domain-relevant queries. For instance, an initial prompt such as \textit{``Does this example belong to the positive class?''} may be refined into a targeted question like \textit{``Does the example mention coughing up blood or fever?''}, better aligned with the semantic distinctions in the data.

\paragraph{Iteration 0 — Initial Generic Prompt}

\textbf{Prompt \( p^{(0)} \):}  
\texttt{Based on the provided context, does this example belong to the positive class? (yes/no)}

\textbf{Gini impurity:} 0.495

\textbf{Group predicted “YES” (sdL):}
\begin{itemize}
    \item \textbf{Correct predictions:} Examples often exhibited severe and systemic symptoms such as coughing blood, weight loss, shortness of breath, rashes, and fatigue.
    \item \textbf{Misclassified “no” cases:} Contained milder or localized symptoms like eye itching, mild sore throat, or isolated skin rashes, which were insufficiently discriminated by the general prompt.
\end{itemize}

\textbf{Group predicted “NO” (sdG):}
\begin{itemize}
    \item \textbf{Correct predictions:} Mostly examples with mild and localized symptoms—itchy nose, minor pain, swelling—without strong systemic indicators.
    \item \textbf{Misclassified “yes” cases:} Actually showed critical signs such as coughing up blood, high fever, significant pain, or weight loss, which the current question failed to detect.
\end{itemize}

\textbf{Global semantic feedback:}  
\emph{“Focusing the question on specific, highly discriminative features—such as the presence of severe systemic symptoms like coughing blood, weight loss, or high fever—can enhance class separation. This targeted approach reduces ambiguity, improves alignment with key features, and minimizes misclassification.”}

\textbf{Revision \(\rightarrow p^{(1)}\):}  
\texttt{Does this example show severe systemic symptoms such as coughing blood, weight loss, or high fever? (yes/no)}

\paragraph{Iteration 1 — Emphasizing a Single Discriminative Feature}

\textbf{Prompt \( p^{(1)} \):}  
\texttt{Does this example show severe systemic symptoms such as coughing blood, weight loss, or high fever? (yes/no)}

\textbf{Gini impurity:} 0.470

\textbf{Group predicted “YES” (sdL):}
\begin{itemize}
    \item \textbf{Correct predictions:} Frequently involved coughing up blood alongside systemic symptoms like nasal congestion, eye itching, fatigue, fever, and skin lesions—indicating strong respiratory or infectious patterns.
    \item \textbf{Misclassified “no” cases:} Contained only mild respiratory issues (e.g., itchy or runny nose) and lacked the critical signal of coughing up blood. The model overgeneralized and flagged weakly indicative cases as positive.
\end{itemize}

\textbf{Group predicted “NO” (sdG):}
\begin{itemize}
    \item \textbf{Correct predictions:} Examples were diverse but lacked specific severe indicators like high fever or respiratory distress. These cases included generalized symptoms (e.g., skin rashes, fatigue) that didn’t clearly indicate the positive class.
    \item \textbf{Misclassified “yes” cases:} Contained classic positive indicators such as high fever, muscle pain, fatigue, and systemic lesions. The model failed to recognize the more complex symptom constellation as positive.
\end{itemize}

\textbf{Global semantic feedback:}  
\emph{“The current question is too broad and includes features (like fever or weight loss) that are insufficiently discriminative. Focus the question on the most specific positive-class indicator — coughing up blood — and remove less reliable features. This should improve class separation by reducing false positives.”}

\textbf{Revision \(\rightarrow p^{(2)}\):}  
\texttt{Does this example show evidence of coughing up blood? (yes/no)}

\paragraph{Iteration 2 — Broadening to Capture Severe Cases}

\textbf{Prompt \( p^{(2)} \):}  
\texttt{Does this example show evidence of coughing up blood? (yes/no)}

\textbf{Gini impurity:} 0.468

\subsection{Prompt Constraints during Prompt Optimization}
\label{appendix:prompt_constraints}

During prompt optimization in ACT, additional constraints are applied to keep questions clear and interpretable while encouraging exploration. The goal of these constraints is twofold: (i) ensure that generated questions remain simple, precise, and auditable, and (ii) incentivize diversity so that ACT explores many different semantic formulations during refinement. The maximum number of logical operators $L$ allowed per question is a controllable hyperparameter in ACT. In all experiments reported in the paper, we set $L = 2$.

\noindent
\colorbox{blue!10}{
\parbox{\dimexpr\linewidth-3\fboxsep\relax}{%
{\bfseries LLM Task: Generate Optimized Question (with Constraints) \par}\smallskip

The following constraints must always be satisfied:\par\smallskip

    -- The question has to be clear and easy. \par
    -- The question must include at least one \textbf{characteristic} different from the previous one. \par
    -- The content of the new question must be significantly different from the current one. \par
    -- Use at most $L$ logical operators (\texttt{and}/\texttt{or}). \par
    -- The question has to be answerable with \texttt{yes} or \texttt{no} only. \par
    -- The question must finish with “(yes/no)?”. \par
    -- Do not use vague words like \texttt{could}, \texttt{might}, or \texttt{possibly}. \par
    -- Do not use blanks or placeholder tags like \texttt{``\string_\string_\string_''} or \texttt{<...>}. \par
    % -- Do not use blanks or placeholder tags like \texttt{"\_\_\_"} or \texttt{<...>}. \par
}}

\subsection{Details on Datasets}
\label{appendix:datasets}

We use five publicly available text classification datasets spanning diverse domains. DIAGNO, SPAM, and JAILBREAK are included because they represent tasks for which structured reasoning is expected to be beneficial, aligning with the types of problems ACT is designed to address. BANKCHURN, a tabular dataset serialized into text, is included because it relies less on pretraining-aligned task semantics and more on learning dataset-specific relationships between attributes and the churn label. IMDB, a widely used benchmark for sentiment classification, is added both to broaden the evaluation and to facilitate comparison with prior work. All datasets are relatively small and mostly balanced, reflecting the low-data prompt-optimization regime in which ACT and comparable methods are typically evaluated~\citep{yuksekgonul2024textgrad}.

\begin{itemize}
    \item \textbf{DIAGNO}~\citep{ninaa510diagnosis}: a medical diagnosis dataset (tuberculosis vs.\ allergic sinusitis) , a derived dataset based on DDXPlus~\cite{fansi2022ddxplus}. We randomly sampled a balanced subset using a fixed random seed, comprising 600 training, 100 validation, and 600 test samples, each with a 50/50 class split between tuberculosis and allergic sinusitis. %We constructed a balanced subset with 600 training, 100 validation, and 600 test samples, ensuring a 50/50 split (300 tuberculosis and 300 allergic sinusitis cases).
    \item %\textbf{SPAM}~\citep{deysi2023spam}: an email spam detection dataset with 600 training, 100 validation, and 600 test samples. 
    \textbf{SPAM}: an email spam detection dataset derived from \citet{deysi2023spam}. 
    We constructed a balanced subset with 600 training, 100 validation, and 600 test samples, following the same procedure as for DIAGNO.
    \item \textbf{JAILBREAK}~\citep{shen2024donowcharacterizingevaluating}: a jailbreak prompt classification dataset with 923 training, 102 validation, and 249 test samples (32 examples have been dropped due to high context %, $>32.696$ 
    tokens).
    
    \item \textbf{BANKCHURN}~\citep{topre2025bankchurn}: a customer–attrition prediction dataset originally provided in tabular form. Following \citet{arzaghi2025intrinsic}, we serialize each record into a short textual profile describing the customer attributes. We constructed a balanced
    subset with 600 training, 100 validation, and 600 test samples, following the same procedure as
    for DIAGNO.

    \item \textbf{IMDB}~\citep{stanford2020imdb}: a sentiment analysis dataset consisting of movie reviews labeled as positive or negative. We constructed a balanced
    subset with 600 training, 100 validation, and 600 test samples, following the same procedure as
    for DIAGNO.

\end{itemize}

\section{Review of Traditional Decision Tree Methods: CART and C4.5}

Classification and Regression trees (CART), proposed by \citet{DBLP:books/wa/BreimanFOS84}, are widely used decision tree algorithms for supervised classification tasks. 
In this work, we focus exclusively on the classification setting. 
%The overall algorithm is outlined in Algorithm~\ref{alg:CART}.

In this subsection we define $\mathcal{D}$ as a tabular dataset \(\mathcal{D} = \{(x_i, y_i)\}_{i=1}^N\), where each instance \(x_i\!\in\!\mathbb{R}^d\) is associated with a target output \(y_i\), CART constructs a binary decision tree by recursively partitioning the dataset based on threshold splits on numeric input features or binary partitions of categorical inputs, selected in a greedy manner. We detail these steps below.

%\paragraph{Node Definition}
%For numerical attributes, CART divides the data into two subsets by selecting an optimal threshold $s$ on a given feature dimension $j$:

%\begin{equation}
%D_L = \{(x, y)\in\mathcal{D}\mid x_j \leq s\}, \quad D_R = \{(x, y)\in\mathcal{D}\mid x_j > s\}.
%\end{equation}
\paragraph{Node Definition.}
At each internal node \( x \) of the tree, we denote by \( \mathcal{D}^x \subseteq \mathcal{D} \) the subset of the training data that reaches node \( x \). In particular, for the root node \( x_0 \), we have \( \mathcal{D}^{x_0} = \mathcal{D} \), the full training dataset.

To split the data at node \( x \), CART evaluates candidate partitions along a feature dimension \( j \) and a threshold \( s \in \mathbb{R} \). The dataset \( \mathcal{D}^x \) is then divided into two subsets:
\begin{equation}
\mathcal{D}_L^x = \{(x_i, y_i) \in \mathcal{D}^x \mid x_{i,j} \leq s\}, \quad
\mathcal{D}_R^x = \{(x_i, y_i) \in \mathcal{D}^x \mid x_{i,j} > s\}.
\end{equation}

These subsets correspond to the training data that will be used at the left and right child nodes of \( x \), respectively. That is, if \( y_L \) and \( y_R \) denote the left and right children of node \( x \), we define \( \mathcal{D}^{y_L} = \mathcal{D}_L^x \) and \( \mathcal{D}^{y_R} = \mathcal{D}_R^x \).

For categorical features, CART evaluates binary partitions over subsets of discrete values, typically by assigning certain categories to the left or right child nodes based on the impurity criterion.

For categorical attributes, CART partitions the data based on whether instances belong or do not belong to specific category subsets.

\paragraph{Best split criterion}
%In classification settings, CART determines the optimal split by minimizing an impurity criterion, typically the Gini impurity. Gini impurity measures how mixed or "impure" a subset is in terms of class labels, aiming for splits that clearly separate classes. It is defined as follows:

%\begin{equation}
%\delta(D_L, D_R) = \frac{|D_L|}{|\mathcal{D}|}\text{Gini}(D_L) + \frac{|D_R|}{|\mathcal{D}|}\text{Gini}(D_R),
%\end{equation}

%where Gini impurity for any subset \(D\) is defined as:

At each node \( x \), CART selects the optimal feature and split threshold by minimizing a node-specific impurity score. This score is typically computed using the Gini impurity, which quantifies the heterogeneity of class labels within a subset.

Given a candidate split of the local dataset \( \mathcal{D}^x \) into \( \mathcal{D}_L^x \) and \( \mathcal{D}_R^x \), the impurity of the split is defined as the weighted sum of the impurities of the two resulting subsets:
\begin{equation}
\delta_j^s = 
\frac{|\mathcal{D}_L^x(j, s)|}{|\mathcal{D}^x|} \cdot \mathrm{Gini}(\mathcal{D}_L^x(j, s)) 
+ \frac{|\mathcal{D}_R^x(j, s)|}{|\mathcal{D}^x|} \cdot \mathrm{Gini}(\mathcal{D}_R^x(j, s)),
\end{equation}

%\begin{equation}
%\mathrm{Gini}(D) = 1 - \sum_{k \in \mathcal{Y}} p_k^2,
%\end{equation}

%where \(p_k\) denotes the proportion of samples belonging to class \(k\) in subset \(D\), and \(\mathcal{Y}\) the set of all possible classes. 

The Gini impurity of any subset \( D \subseteq \mathcal{D} \) is given by:
\begin{equation}
\mathrm{Gini}(D) = 1 - \sum_{k \in \mathcal{Y}} p_k^2,
\end{equation}
where \( \mathcal{Y} \) denotes the set of possible class labels, and \( p_k \) is the empirical proportion of class \( k \) in the dataset \( D \).

A Gini impurity of 0 indicates perfect purity—i.e., all instances in the subset belong to a single class—while a value closer to 0.5 (in binary classification) indicates a highly mixed, impure subset.

\paragraph{Best split search}
The CART algorithm constructs the decision tree recursively using a greedy strategy: at each internal node \(x\), the best split is selected by optimizing a local impurity criterion over the current subset \(\mathcal{D}^x\). This split is chosen by evaluating its effect on the resulting child nodes—i.e., by computing the weighted impurity of the partition it induces. In practice, this local optimization is typically performed via exhaustive search over all candidate features and thresholds, though heuristic approximations may be used in high-dimensional settings.

The recursive construction proceeds until a stopping criterion is met, such as achieving node purity, reaching a minimum number of samples, or exceeding a predefined maximum depth \(d_{\max}\).

\paragraph{Tree construction.}
Decision trees are constructed recursively. At each node, the split minimizing impurity is applied to partition the data, and the procedure recurses on the resulting subsets. Recursion terminates when purity is reached or a predefined stopping condition is met (e.g., maximum depth, minimum node size), and the node is labeled with the majority class.

\paragraph{C4.5.}  
C4.5~\citep{quinlan1993c4} follows the same recursive tree-building process as CART but differs mainly in two respects: 
(i) it chooses splits by maximizing the \emph{information gain ratio} rather than minimizing Gini impurity, and 
(ii) it allows multiway splits on categorical attributes, whereas CART restricts all splits to be binary.

While both CART and C4.5 have become canonical algorithms for structured data, they are not directly applicable to unstructured domains such as text or images. This limitation motivates our proposed Agentic Classification Tree (ACT).

\section{Additional Qualitative Results}

\subsection{Qualitative analysis of the questions generated by ACT}
\label{sec:comparison-experts}

\begin{table}[ht]
    \centering
    \caption{Symptoms comparison between ACT and medical sources.}
    \begin{tabular}{c|c|c}
       Questions in common  & ACT only & Medical sources only  \\
       \midrule
        Coughing up blood & Shortness of breath & Chest pain\\
        Fever &  Severe pain & Headache \\
        Weight loss &  &  Increased sweating\\
        Swollen lymph nodes &  & Loss of appetite \\
        Fatigue & & \\
        % & & \\
        %Skin rashes & & \\
        %Headache & & \\
    \end{tabular}
    \label{qualitative-comparison-questions}
\end{table}

In addition to its tree structure, another crucial element ensuring the interpretability of ACT lies in how relevant the generated questions are. 
For this purpose, we propose to compare these questions with the ones that domain experts would have asked, focusing on the DIAGNO dataset. 
We survey several relevant medical sources to collect a list of symptoms that are commonly used for tuberculosis diagnosis. These sources include websites of the World Health Organization\footnote{https://www.who.int/news-room/fact-sheets/detail/tuberculosis} and the British National Health Service\footnote{https://www.nhs.uk/conditions/tuberculosis-tb/}, and a literature review on tuberculosis diagnosis~\cite{storla2008systematic}.

Table~\ref{qualitative-comparison-questions} shows a side-by-side comparison of the symptoms identified by ACT as positively associated with tuberculosis, and those most commonly identified in the aforementioned medical sources, differentiating the symptoms identified by both ACT and the medical sources (left column), from those that were exclusive (center and right columns). We see that most of the symptoms appear in common. Some of the differences may be explained by ambiguity between some symptoms (e.g. "severe pain" vs. chest pain and headaches). Others may come from the binary classification setting considered, opposing tuberculosis cases to allergy, potentially resulting in some symptoms (e.g. loss of appetite) not being represented in the dataset.  Overall, this tends to show that ACT is indeed able to identify relevant symptoms to make its predictions, showcasing its utility and reliability.

\subsection{Agentic Classification Tree for SPAM and JAILBREAK Dataset}
\label{appendix:act_spam_jailbreak}

\begin{figure}[ht]
\centering
% ---- Decision Tree ----
\begin{minipage}{0.95\textwidth}
    \centering
    \resizebox{\textwidth}{!}{%
    \begin{forest}
    for tree={
        align=center,
        parent anchor=south,
        child anchor=north,
        l sep+=12pt,
        s sep+=6pt,
        draw,
        rounded corners=4pt,
        minimum width=2.4cm,
        minimum height=0.8cm,
        font=\footnotesize,
        edge={->, >=latex, thick},
    }
    [{Does the text contain promotional\\language or marketing terms?}, fill=blue!10
        [{Does this text use excessive exclamation\\points, grammatical errors, or\\poor writing quality?}, edge label={node[midway, font=\scriptsize, fill=white, inner sep=1pt]{yes}}, fill=blue!10
            [{\textbf{SPAM}}, edge label={node[midway, font=\scriptsize, fill=white, inner sep=1pt]{yes}}, fill=red!30, draw=red!70!black, thick]
            [{Does this example primarily aim to\\generate immediate engagement\\(likes, comments, shares)?}, edge label={node[midway, font=\scriptsize, fill=white, inner sep=1pt]{no}}, fill=blue!10
                [{\textbf{SPAM}}, edge label={node[midway, font=\scriptsize, fill=white, inner sep=1pt]{yes}}, fill=red!30, draw=red!70!black, thick]
                [{\textbf{HAM}}, edge label={node[midway, font=\scriptsize, fill=white, inner sep=1pt]{no}}, fill=green!30, draw=green!70!black, thick]
            ]
        ]
        [{Does this text rely heavily on\\emotionally charged language\\or sensational claims?}, edge label={node[midway, font=\scriptsize, fill=white, inner sep=1pt]{no}}, fill=blue!10
            [{\textbf{SPAM}}, edge label={node[midway, font=\scriptsize, fill=white, inner sep=1pt]{yes}}, fill=red!30, draw=red!70!black, thick]
            [{Does this example aim to generate\\immediate revenue or sales?}, edge label={node[midway, font=\scriptsize, fill=white, inner sep=1pt]{no}}, fill=blue!10
                [{\textbf{SPAM}}, edge label={node[midway, font=\scriptsize, fill=white, inner sep=1pt]{yes}}, fill=red!30, draw=red!70!black, thick]
                [{\textbf{HAM}}, edge label={node[midway, font=\scriptsize, fill=white, inner sep=1pt]{no}}, fill=green!30, draw=green!70!black, thick]
            ]
        ]
    ]
    \end{forest}
    }
\end{minipage}
\caption{Decision tree of depth 3 generated by the Gemma3-4B model for spam email classification. The tree distinguishes between spam and legitimate email (ham) through hierarchical semantic questions about promotional content, writing quality, and intent. Each internal node represents a binary question optimized through the ACT framework to maximize class separation, with leaf nodes indicating the final classification based on the majority class of training examples.}
\label{fig:spam_tree_qwen3}
\end{figure}

% ---- Decision Tree ----
\begin{figure}[ht]
\centering
\begin{minipage}{0.95\textwidth}
    \centering
    \resizebox{\textwidth}{!}{%
    \begin{forest}
    for tree={
        align=center,
        parent anchor=south,
        child anchor=north,
        l sep+=12pt,
        s sep+=6pt,
        draw,
        rounded corners=4pt,
        minimum width=2.4cm,
        minimum height=0.8cm,
        font=\footnotesize,
        edge={->, >=latex, thick},
    }
    [{Does this text explicitly instruct an AI to\\bypass ethical, legal, or moral constraints?}, fill=blue!10
        [{Does the prompt implicitly or explicitly instruct\\the AI to override policy or safety restrictions?}, edge label={node[midway, font=\scriptsize, fill=white, inner sep=1pt]{yes}}, fill=blue!10
            [{\textbf{JB}}, edge label={node[midway, font=\scriptsize, fill=white, inner sep=1pt]{yes}}, fill=red!30, draw=red!70!black, thick]
            [{Is this text a direct instruction to generate\\harmful, unethical, or illegal content in\\a role-playing or creative context?}, edge label={node[midway, font=\scriptsize, fill=white, inner sep=1pt]{no}}, fill=blue!10
                [{\textbf{JB}}, edge label={node[midway, font=\scriptsize, fill=white, inner sep=1pt]{yes}}, fill=red!30, draw=red!70!black, thick]
                [{\textbf{Not JB}}, edge label={node[midway, font=\scriptsize, fill=white, inner sep=1pt]{no}}, fill=green!30, draw=green!70!black, thick]
            ]
        ]
        [{Does the example require imaginative or\\persona-driven instructions that guide the\\response toward creative or roleplay content?}, edge label={node[midway, font=\scriptsize, fill=white, inner sep=1pt]{no}}, fill=blue!10
            [{Does this roleplay scenario explicitly request\\or involve role-playing as an AI (e.g., 'I am DAN',\\'I am JailBreak') that ignores ethical,\\legal, or moral guidelines?}, edge label={node[midway, font=\scriptsize, fill=white, inner sep=1pt]{yes}}, fill=blue!10
                [{\textbf{JB}}, edge label={node[midway, font=\scriptsize, fill=white, inner sep=1pt]{yes}}, fill=red!30, draw=red!70!black, thick]
                [{\textbf{Not JB}}, edge label={node[midway, font=\scriptsize, fill=white, inner sep=1pt]{no}}, fill=green!30, draw=green!70!black, thick]                
            ]
            [{Is this text a neutral, educational, or\\role-playing scenario with fictional characters\\in a non-offensive context?}, edge label={node[midway, font=\scriptsize, fill=white, inner sep=1pt]{no}}, fill=blue!10
                [{\textbf{JB}}, edge label={node[midway, font=\scriptsize, fill=white, inner sep=1pt]{no}}, fill=red!30, draw=red!70!black, thick]
                [{\textbf{Not JB}}, edge label={node[midway, font=\scriptsize, fill=white, inner sep=1pt]{yes}}, fill=green!30, draw=green!70!black, thick]
            ]
        ]
    ]
    \end{forest}
    }
\end{minipage}
\caption{Decision tree of depth 3 generated by the Qwen3-4b model for jailbreak prompt classification. The tree recursively partitions inputs through binary natural-language questions optimized to distinguish between jailbreak attempts (JB) and legitimate prompts (Not JB). Each internal node contains a semantically meaningful question discovered through iterative prompt refinement, with terminal nodes indicating the final classification based on the majority class of training examples reaching that leaf.}
\label{fig:jailbreak_tree_qwen3}
\end{figure}

\subsection{Qualitative comparison with TF-IDF + CART}
\label{sec:comparison-tfidf}

In this section we aim to illustrate the benefits of leveraging LLM agents to directly deal with unstructured data by comparing ACT with the traditional TF-IDF and CART combination. The tree resulting from this procedure for the DIAGNO dataset is shown in Figure~\ref{fig:tree-tfidf}, and in Figure~\ref{fig:tree-tfidf-spam} for the SPAM dataset. In both cases, we observe the limitation of this approach, as nodes built only on one word provide little help in understanding the model. Worse, these words may often end up being generic adverbs and prepositions (e.g. "our", "any"), resulting ultimately in explanations that are both hard to understand and leverage. On the other hand, the decision trees learned by ACT for these datasets can be easily understood, and verified, by a user.

\begin{figure}[ht]
\centering
\begin{minipage}{0.95\textwidth}
    \centering
    \resizebox{\textwidth}{!}{%
    \begin{forest}
    for tree={
        align=center,
        parent anchor=south,
        child anchor=north,
        l sep+=12pt,
        s sep+=6pt,
        draw,
        rounded corners=4pt,
        minimum width=2.4cm,
        minimum height=0.8cm,
        font=\footnotesize,
        edge={->, >=latex, thick},
    }
    [{nose $\leq 0.074$}, fill=blue!10
        [{itching $\leq 0.052$}, edge label={node[midway, font=\scriptsize, fill=white, inner sep=1pt]{yes}}, fill=blue!10
            [{nausea $\leq 0.192$}, edge label={node[midway, font=\scriptsize, fill=white, inner sep=1pt]{yes}}, fill=blue!10[{\textbf{TB}}, edge label={node[midway, font=\scriptsize, fill=white, inner sep=1pt]{yes}}, fill=red!30, draw=red!70!black, thick]
                [{\textbf{Not TB}}, edge label={node[midway, font=\scriptsize, fill=white, inner sep=1pt]{no}}, fill=green!30, draw=green!70!black, thick]
            %draw=red!70!black, thick
            ]
            [{also $\leq 0.294$}, edge label={node[midway, font=\scriptsize, fill=white, inner sep=1pt]{no}}, fill=blue!10
                [{\textbf{Not TB}}, edge label={node[midway, font=\scriptsize, fill=white, inner sep=1pt]{no}}, fill=green!30, draw=green!70!black, thick]
                [{\textbf{TB}}, edge label={node[midway, font=\scriptsize, fill=white, inner sep=1pt]{yes}}, fill=red!30, draw=red!70!black, thick]
            ]
        ]
        [{nasal $\leq 0.238$}, edge label={node[midway, font=\scriptsize, fill=white, inner sep=1pt]{no}}, fill=blue!10
            [{constant $\leq 0.289$}, edge label={node[midway, font=\scriptsize, fill=white, inner sep=1pt]{yes}}, fill=blue!10
                [{\textbf{Not TB}}, edge label={node[midway, font=\scriptsize, fill=white, inner sep=1pt]{no}}, fill=green!30, draw=green!70!black, thick][{\textbf{TB}}, edge label={node[midway, font=\scriptsize, fill=white, inner sep=1pt]{yes}}, fill=red!30, draw=red!70!black, thick]              
            ]
            [{nose $\leq$ 0.309}, edge label={node[midway, font=\scriptsize, fill=white, inner sep=1pt]{no}}, fill=blue!10
                [{\textbf{TB}}, edge label={node[midway, font=\scriptsize, fill=white, inner sep=1pt]{no}}, fill=red!30, draw=red!70!black, thick]
                [{\textbf{Not TB}}, edge label={node[midway, font=\scriptsize, fill=white, inner sep=1pt]{yes}}, fill=green!30, draw=green!70!black, thick]
            ]
        ]
    ]
    \end{forest}
    }
\end{minipage}
\caption{Decision tree of depth 3 generated by training a CART model after performing a TF-IDF preprocessing step on the DIAGNO dataset.}
\label{fig:tree-tfidf}
\end{figure}

\begin{figure}[ht]
\centering
\begin{minipage}{0.95\textwidth}
    \centering
    \resizebox{\textwidth}{!}{%
    \begin{forest}
    for tree={
        align=center,
        parent anchor=south,
        child anchor=north,
        l sep+=12pt,
        s sep+=6pt,
        draw,
        rounded corners=4pt,
        minimum width=2.4cm,
        minimum height=0.8cm,
        font=\footnotesize,
        edge={->, >=latex, thick},
    }
    [{our $\leq 0.041$}, fill=blue!10
        [{your $\leq 0.06$}, edge label={node[midway, font=\scriptsize, fill=white, inner sep=1pt]{yes}}, fill=blue!10
            [{now $\leq 0.055$}, edge label={node[midway, font=\scriptsize, fill=white, inner sep=1pt]{yes}}, fill=blue!10
                [{\textbf{HAM}}, edge label={node[midway, font=\scriptsize, fill=white, inner sep=1pt]{yes}}, fill=green!30, draw=green!70!black, thick]
                [{\textbf{SPAM}}, edge label={node[midway, font=\scriptsize, fill=white, inner sep=1pt]{no}}, fill=red!30, draw=red!70!black, thick]
            ]
            [{any $\leq 0.038$}, edge label={node[midway, font=\scriptsize, fill=white, inner sep=1pt]{no}}, fill=blue!10
                [{\textbf{SPAM}}, edge label={node[midway, font=\scriptsize, fill=white, inner sep=1pt]{yes}}, fill=red!30, draw=red!70!black, thick]
                [{\textbf{HAM}}, edge label={node[midway, font=\scriptsize, fill=white, inner sep=1pt]{no}}, fill=green!30, draw=green!70!black, thick]
            ]
        ]
        [{announce $\leq 0.1$}, edge label={node[midway, font=\scriptsize, fill=white, inner sep=1pt]{no}}, fill=blue!10
            [{severe $\leq 0.13$}, edge label={node[midway, font=\scriptsize, fill=white, inner sep=1pt]{yes}}, fill=blue!10
                [{\textbf{SPAM}}, edge label={node[midway, font=\scriptsize, fill=white, inner sep=1pt]{yes}}, fill=red!30, draw=red!70!black, thick]
                [{\textbf{HAM}}, edge label={node[midway, font=\scriptsize, fill=white, inner sep=1pt]{no}}, fill=green!30, draw=green!70!black, thick]
            ]
            [{\textbf{HAM}}, edge label={node[midway, font=\scriptsize, fill=white, inner sep=1pt]{no}}, fill=green!30, draw=green!70!black, thick]
        ]
    ]
    \end{forest}
    }
\end{minipage}
\caption{Decision tree of depth 3 generated by training a CART model after performing a TF-IDF preprocessing step on the SPAM dataset.}
\label{fig:tree-tfidf-spam}
\end{figure}

\section{Additional Results and Baseline Implementation details}
\label{appendix:classical_baselines}

\subsection{Additional Baselines and ACT Configuration Results}

To complement the results shown in Table~\ref{tab:results_acart} we expand this table in Table~\ref{tab:results_acart_full_part_1} and Table~\ref{tab:results_acart_full_part_2}, which include additional results for the ACT $(d=3, k=20)$ and $(d=4, k=10)$ configurations, as well as strong but not interpretable baselines:

\begin{itemize}
    \item \textbf{TF-IDF + XGBoost}: a gradient-boosted tree ensemble trained on TF-IDF representations of the texts~\citep{chen2016xgboost}. We tune both the maximum tree depth and the number of trees (see Appendix~\ref{appendix:classical_baselines}), and report the best-performing configuration per dataset.

    \item \textbf{Fine-tuned BERT and RoBERTa classifiers}: supervised Transformer baselines initialized from pretrained BERT and RoBERTa encoders~\citep{devlin2019bert,liu2019roberta} and fine-tuned end-to-end on each dataset for text classification. These models provide strong predictive performance references, but do not yield explicit, human-readable decision rules.
\end{itemize}

\begin{table}[htbp!]
    \centering
    \small
    \caption{Comparison of classification methods across the first three datasets and four LLMs. Results show accuracy and F1-score on the test set (in \%) for nine benchmarks, as well as the proposed ACT with varying hyperparameters ($d$: tree depth, $k$: optimization steps per node).}
    \label{tab:results_acart_full_part_1}
    
    \resizebox{0.85\textwidth}{!}{%
    \begin{tabular}{l rr|rr|rr|rr|rr}
    \toprule
    \multirow{2}{*}{\textbf{Method}}
    & \multicolumn{2}{c}{\textbf{Gemma3 4b}}
    & \multicolumn{2}{c}{\textbf{GPT-4.1 Nano}}
    & \multicolumn{2}{c}{\textbf{GPT-4.1 Mini}}
    & \multicolumn{2}{c}{\textbf{Qwen3 4b}}
    & \multicolumn{2}{c}{\textbf{Avg}} \\
    \cmidrule(lr){2-3} \cmidrule(lr){4-5} \cmidrule(lr){6-7} \cmidrule(lr){8-9} \cmidrule(lr){10-11}
    & \textbf{Acc} & \textbf{F1}
    & \textbf{Acc} & \textbf{F1}
    & \textbf{Acc} & \textbf{F1}
    & \textbf{Acc} & \textbf{F1}
    & \textbf{Acc} & \textbf{F1} \\
    \midrule
    
    % ===================== DIAGNO =====================
    % \multicolumn{11}{l}{\textbf{DIAGNO}} \\
    \textbf{DIAGNO} & & & & & & & & & & \\
    CoT (0-shot) & 61.5 & 69.5 & 63.3 & 54.8 & 61.2 & 51.4 & 63.2 & 41.7 & 62.3 & 54.4 \\
    DSPy (BFSR, 8 demos) & 64.0 & 68.2 & 68.7 & 61.2 & 64.5 & 59.4 & 63.3 & 54.9 & 64.0 & 60.9 \\
    \textsc{TextGrad} & 63.3 & 56.7 & 64.5 & 53.0 & 65.8 & 55.6 & 64.0 & 60.0 & 64.4 & 56.3 \\
    TF-IDF + CART ($d{=}3$) & $--$ & $--$ & $--$ & $--$ & $--$ & $--$ & $--$ & $--$ & $78.8$ & $78.9$ \\
    TF-IDF + XGBoost ($d{=}4, nt{=}200 $ ) 
    & $--$ & $--$ & $--$ & $--$ & $--$ & $--$ & $--$ & $--$ & $83.0$ & $83.5$\\
    BERT & $--$ & $--$ & $--$ & $--$ & $--$ & $--$ & $--$ & $--$ & $90.3$ & $91.1$\\
    RoBERTa & $--$ & $--$ & $--$ & $--$ & $--$ & $--$ & $--$ & $--$ & $89.8$ & $90.4$\\
    Con. tag. & $65.5$ & $63.9$ & $59.3$ & $58.1$ & $75.3$ & $78.6$ & $58.8$ & $64.7$ & $64.7$ & $66.3$ \\
    Rule Fit & $--$ & $--$ & $--$ & $--$ & $--$ & $--$ & $--$ & $--$ & $80.2$ & $79.3$ \\
    \textsc{ACT} ($d{=}3$, $k{=}10$) & 65.3 & 66.4 & 66.8 & 67.0 & 77.3 & 75.2 & 64.8 & 61.3 & 68.6 & 67.7 \\
    \textsc{ACT} ($d{=}3$, $k{=}20$) & $67.0$ & $67.9$ & $67.2$ & $69.0$ & $76.2$ & $76.9$ & $65.5$ & $61.8$ & $69.0$ & $68.9$\\
    \textsc{ACT} ($d{=}4$, $k{=}10$) & $66.3$ & $\textbf{71.1}$ & $\textbf{70.5}$ & $70.1$ & $74.3$ & $70.1$ & $66.7$ & $66.8$ & $69.5$ & $69.5$\\
    \textsc{ACT} ($d{=}4$, $k{=}20$) & \textbf{68.3} & 70.8 & 70.3 & \textbf{70.8} & \textbf{82.3} & \textbf{81.8} & \textbf{70.3} & \textbf{73.5} & 72.8 & 74.2 \\
    
    \midrule
    
    % ===================== SPAM =====================
    %\multicolumn{11}{l}{\textbf{SPAM}} \\
    \textbf{SPAM} & & & & & & & & & & \\
    CoT (0-shot) & 73.3 & 78.6 & 95.5 & 95.3 & 95.7 & 95.5 & 93.2 & 93.6 & 89.4 & 90.8 \\
    DSPy (BFSR, 8 demos) & 95.0 & 95.2 & 98.3 & 98.3 & 98.2 & 98.2 & 97.2 & 97.2 & 97.2 & 97.2 \\
    \textsc{TextGrad} & 95.0 & 94.7 & 97.3 & 97.3 & 96.7 & 96.6 & 96.0 & 96.0 & 96.3 & 96.2 \\ 
    TF-IDF + CART ($d{=}5$) & $--$ & $--$ & $--$ & $--$ & $--$ & $--$ & $--$ & $--$ & $92.7$ & $93.1$ \\
    TF-IDF + XGBoost ($d{=}3, nt{=}200 $ ) 
    & $--$ & $--$ & $--$ & $--$ & $--$ & $--$ & $--$ & $--$ & $96.3$ & $96.3$\\
    BERT & $--$ & $--$ & $--$ & $--$ & $--$ & $--$ & $--$ & $--$ & $99.5$ & $99.5$\\
    RoBERTa & $--$ & $--$ & $--$ & $--$ & $--$ & $--$ & $--$ & $--$ & $99.3$ & $99.3$\\
    Con. tag. & $83.2$ & $85.5$ & $64.2$ & $45.8$ & $82.3$ & $85.0$ & $62.3$ & $39.6$ & $73.0$ & $64.0$ \\
    Rule Fit & $--$ & $--$ & $--$ & $--$ & $--$ & $--$ & $--$ & $--$ & $97.0$ & $96.9$ \\
    \textsc{ACT} ($d{=}2$, $k{=}5$) & 95.8 & 95.8 & 98.7 & 98.6 & 98.2 & 98.2 & 96.7 & 96.6 & 97.4 & 97.3 \\
    \textsc{ACT} ($d{=}2$, $k{=}10$) & $96.5$ & $96.6$ & $98.8$ & $98.8$ & $99.2$ & $99.2$ & $98.5$ & $98.5$ & $98.4$ & $98.4$ \\
    \textsc{ACT} ($d{=}3$, $k{=}5$) & $98.2$ & $98.1$ & $97.0$ & $96.9$ & $99.0$ & $99.0$ & $98.5$ & $98.5$ & $98.3$ & $98.3$\\
    \textsc{ACT} ($d{=}3$, $k{=}10$) & \textbf{98.5} & \textbf{98.5} & \textbf{99.2} & \textbf{99.2} & \textbf{99.5} & \textbf{99.5} & \textbf{98.8} & \textbf{98.8} & 99.0 & 99.0 \\
    
    \midrule
    
    % ===================== JAILBREAK =====================
    %\multicolumn{11}{l}{\textbf{JAILBREAK}} \\
    \textbf{JAILBREAK} & & & & & & & & & & \\
    
    CoT (0-shot) & 77.9 & 82.2 & 85.5 & 83.5 & 93.6 & 93.3 & 81.1 & 77.7 & 84.5 & 84.2 \\
    DSPy (BFSR, 8 demos) & 89.6 & 89.3 & 91.6 & \textbf{91.6} & 94.8 & 94.8 & 85.5 & 83.9 & 90.4 & 89.9 \\
    \textsc{TextGrad} & \textbf{91.6} & \textbf{91.3} & 90.4 & 90.3 & 94.8 & 95.1 & 88.8 & 90.0 & 91.4 & 91.7 \\
    TF-IDF + CART ($d{=}4$) & $--$ & $--$ & $--$ & $--$ & $--$ & $--$ & $--$ & $--$ & $92.8$ & $92.4$\\
    TF-IDF + XGBoost ($d{=}2, nt{=}200 $ ) 
    & $--$ & $--$ & $--$ & $--$ & $--$ & $--$ & $--$ & $--$ & $96.8$ & $96.8$\\
    BERT & $--$ & $--$ & $--$ & $--$ & $--$ & $--$ & $--$ & $--$ & $98.4$ & $98.4$\\
    RoBERTa & $--$ & $--$ & $--$ & $--$ & $--$ & $--$ & $--$ & $--$ & $99.6$ & $99.6$ \\
    Con. tag. & $82.7$ & $82.2$ & $58.6$ & $31.8$ & $82.7$ & $79.8$ & $76.3$ & $70.1$ & $75.1$ & $66.0$ \\
    Rule Fit & $--$ & $--$ & $--$ & $--$ & $--$ & $--$ & $--$ & $--$ & $96.8$ & $96.8$ \\
    \textsc{ACT} ($d{=}3$, $k{=}10$) & 82.3 & 82.7 & 85.5 & 85.1 & 95.2 & 95.5 & 82.7 & 82.6 & 86.4 & 86.5 \\
    \textsc{ACT} ($d{=}3$, $k{=}20$) & $84.3$ & $86.5$ & $91.2$ & $90.5$ & $96.2$ & $96.6$ & $87.6$ & $86.6$ & $89.8$ & $90.1$ \\
    \textsc{ACT} ($d{=}4$, $k{=}10$) & $90.0$ & $90.1$ & $91.6$ & $91.0$ & $97.2$ & $97.2$ & $85.1$ & $83.9$ & $91.0$ & $90.6$ \\
    \textsc{ACT} ($d{=}4$, $k{=}20$) & 90.8 & 90.0 & \textbf{92.0} & 91.5 & \textbf{98.8} & \textbf{98.8} &\textbf{92.0} & \textbf{92.7} & 93.4 & 93.3 \\
    \bottomrule
    \end{tabular}%
    }
\end{table}

\begin{table}[htbp!]
\centering
\small
\caption{Comparison of classification methods across the two remaining datasets and four LLMs. Results show accuracy and F1-score on the test set (in \%) for nine benchmarks, as well as the proposed ACT with varying hyperparameters ($d$: tree depth, $k$: optimization steps per node).}
\label{tab:results_acart_full_part_2}

\resizebox{0.85\textwidth}{!}{%
\begin{tabular}{l rr|rr|rr|rr|rr}
\toprule
\multirow{2}{*}{\textbf{Method}}
& \multicolumn{2}{c}{\textbf{Gemma3 4b}}
& \multicolumn{2}{c}{\textbf{GPT-4.1 Nano}}
& \multicolumn{2}{c}{\textbf{GPT-4.1 Mini}}
& \multicolumn{2}{c}{\textbf{Qwen3 4b}}
& \multicolumn{2}{c}{\textbf{Avg}} \\
\cmidrule(lr){2-3} \cmidrule(lr){4-5} \cmidrule(lr){6-7} \cmidrule(lr){8-9} \cmidrule(lr){10-11}
& \textbf{Acc} & \textbf{F1}
& \textbf{Acc} & \textbf{F1}
& \textbf{Acc} & \textbf{F1}
& \textbf{Acc} & \textbf{F1}
& \textbf{Acc} & \textbf{F1} \\
\midrule

% ===================== BANKCHURN =====================
%\multicolumn{11}{l}{\textbf{BANKCHURN}} \\
\textbf{BANKCHURN} & & & & & & & & & & \\
CoT (0-shot) & 47.8 & 48.1 & 48.5 & 25.9 & 52.8 & 60.5 & 50.2 & 37.3 & 49.8 & 43.0 \\
DSPy (BFSR, 8 demos) & 52.8 & 55.6 & 52.3 & 42.8 & 57.2 & 58.5 & 54.0 & 52.2 & 54.1 & 52.3 \\
\textsc{TextGrad} & 52.0 & 60.0 & 53.3 & 60.3 & 55.3 & 61.0 & 52.8 & 61.4 & 53.4 & 60.7 \\
TF-IDF + CART ($d{=}5$) & $--$ & $--$ & $--$ & $--$ & $--$ & $--$ & $--$ & $--$ & $63.2$ & $66.8$\\
TF-IDF + XGBoost ($d{=}2, nt{=}200 $ ) 
& $--$ & $--$ & $--$ & $--$ & $--$ & $--$ & $--$ & $--$ & $56.7$ & $48.8$\\
BERT & $--$ & $--$ & $--$ & $--$ & $--$ & $--$ & $--$ & $--$ & $71.7$ & $71.6$\\
RoBERTa & $--$ & $--$ & $--$ & $--$ & $--$ & $--$ & $--$ & $--$ & $72.3$ & $72.3$ \\
Con. tag. & $57.2$ & $60.2$ & $52.0$ & $59.8$ & $59.7$ & $60.8$ & $54.3$ & $61.6$ & $55.8$ & $60.6$ \\
Rule Fit & $--$ & $--$ & $--$ & $--$ & $--$ & $--$ & $--$ & $--$ & $63.0$ & $59.8$ \\
\textsc{ACT} ($d{=}3$, $k{=}10$) & 56.3 & 42.3 & 58.8 & 67.4 & 60.5 & 64.6 & 57.5 & 62.7 & 58.3 & 59.3 \\
\textsc{ACT} ($d{=}3$, $k{=}20$) & $58.0$ & $47.8$ & $\textbf{67.0}$ & $67.5$ & $\textbf{69.2}$ & $65.3$ & $\textbf{62.7}$ & $\textbf{63.7}$ & $64.2$ & $61.1$ \\
\textsc{ACT} ($d{=}4$, $k{=}10$) & $61.7$ & $65.3$ & $61.7$ & $60.9$ & $69.0$ & $65.0$ & $60.9$ & $60.2$ & $63.3$ & $62.9$ \\
\textsc{ACT} ($d{=}4$, $k{=}20$) & \textbf{65.2} & \textbf{70.7} & 65.0 & \textbf{69.1} & 68.7 & \textbf{68.6} & 62.0 & 60.0 & 65.2 & 67.1 \\

\midrule

% ===================== IMBD =====================
%\multicolumn{11}{l}{\textbf{IMDB}} \\
\textbf{IMDB} & & & & & & & & & & \\
CoT (0-shot) & 92.3 & 91.9 & 94.2 & 94.1 & 95.7 & 95.6 & 94.2 & 94.0 & 94.1 & 93.9 \\
DSPy (BFSR, 8 demos) & \textbf{93.7} & \textbf{93.6} & \textbf{95.5} & \textbf{95.5} & \textbf{96.8} & \textbf{96.8} & 94.8 & 94.8 & 95.2 & 95.2 \\
\textsc{TextGrad} & 93.2 & 92.9 & 95.2 & 95.1 & 95.8 & 95.8 & 94.7 & 94.5 & 94.7 & 94.6 \\
TF-IDF + CART ($d{=}3$) & $--$ & $--$ & $--$ & $--$ & $--$ & $--$ & $--$ & $--$ & $66.8$ & $72.8$\\
TF-IDF + XGBoost ($d{=}2, nt{=}200 $ ) 
& $--$ & $--$ & $--$ & $--$ & $--$ & $--$ & $--$ & $--$ & $76.2$ & $76.7$\\
BERT & $--$ & $--$ & $--$ & $--$ & $--$ & $--$ & $--$ & $--$ & $90.2$ & $90.1$\\
RoBERTa & $--$ & $--$ & $--$ & $--$ & $--$ & $--$ & $--$ & $--$ & $92.7$ & $92.7$ \\
Con. tag. & $74.8$ & $76.9$ & $56.2$ & $31.7$ & $50.0$ & $66.7$ & $51.3$ & $63.8$ & $58.1$ & $59.8$ \\
Rule Fit & $--$ & $--$ & $--$ & $--$ & $--$ & $--$ & $--$ & $--$ & $75.2$ & $73.7$ \\
\textsc{ACT} ($d{=}3$, $k{=}10$) & 93.2 & 93.2 & 92.8 & 92.8 & 95.7 & 95.7 & 93.8 & 93.9 & 93.9 & 93.9 \\
\textsc{ACT} ($d{=}3$, $k{=}20$) & \textbf{93.7} & \textbf{93.6} & 94.7 & 94.7 & 96.3 & 96.2 & \textbf{95.5} & \textbf{95.5} & 95.1 & 95.0 \\

\bottomrule
\end{tabular}%
}
\end{table}

\subsection{Full Tree-Based Baselines Results}
To complement the prompting-based baselines used in the main paper, we report in Table~\ref{tab:cart_xgb_full} an extended comparison with purely symbolic lexical models. These include TF--IDF + CART with depths ranging from 2 to 5, and TF--IDF + XGBoost with multiple depth and tree-count configurations. These models provide a transparent, non-LLM control group and clarify the performance gap between ACT and traditional supervised baselines across all five datasets. As shown below, ACT remains competitive to these methods while offering semantically grounded, human-readable decision rules.

\begin{table}[ht!]
\setlength{\aboverulesep}{0pt}
\setlength{\belowrulesep}{0pt}
\setlength{\cmidrulekern}{0pt}
\centering
\caption{Performance of TF-IDF + CART and TF-IDF + XGBoost across all datasets.}
\label{tab:cart_xgb_full}
\resizebox{\textwidth}{!}{
\begin{tabular}{lcccc|cccccc}
\toprule
\textbf{Dataset / Split} 
& \textbf{CART d=2} & \textbf{CART d=3} & \textbf{CART d=4} & \textbf{CART d=5}
& \textbf{XGB (2,50)} & \textbf{XGB (2,200)} & \textbf{XGB (3,50)} & \textbf{XGB (3,200)}
& \textbf{XGB (4,50)} & \textbf{XGB (4,200)} \\
\midrule

\textbf{DIAGNO – Train} 
& 82.0 & 84.3 & 86.5 & \textbf{88.8} 
& 86.2 & 94.7 & 89.2 & 98.7 & 93.5 & \textbf{99.8} \\

\textbf{DIAGNO – Test} 
& 77.5 & \textbf{78.8} & 73.7 & 76.3 
& 81.0 & 81.5 & 81.8 & 81.7 & 82.2 & \textbf{83.0} \\

\textbf{SPAM – Train} 
& 87.8 & 92.3 & 93.5 & \textbf{94.7} 
& 97.8 & 99.7 & 98.7 & 99.8 & 99.2 & \textbf{100.0} \\

\textbf{SPAM – Test} 
& 85.8 & 89.3 & 90.7 & \textbf{92.7} 
& 94.3 & 96.2 & 93.8 & \textbf{96.3} & 94.5 & 96.0 \\

\textbf{JAILBREAK – Train} 
& 88.3 & 91.1 & 93.1 & \textbf{94.3} 
& 95.6 & 97.9 & 96.5 & 98.5 & 97.1 & \textbf{99.0} \\

\textbf{JAILBREAK – Test} 
& 86.4 & 92.4 & \textbf{92.8} & 91.9 
& 95.2 & \textbf{96.8} & 95.2 & 96.4 & 95.2 & \textbf{96.8} \\

\textbf{BANKCHURN – Train} 
& 59.7 & 63.0 & 63.7 & \textbf{66.7} 
& 71.5 & 84.5 & 79.3 & 90.0 & 84.2 & \textbf{93.2} \\

\textbf{BANKCHURN – Test} 
& 62.7 & 62.8 & 62.8 & \textbf{63.2} 
& 53.2 & \textbf{56.7} & 54.3 & 53.7 & 54.8 & 55.3 \\

\textbf{IMDB – Train} 
& 68.2 & 71.0 & 71.8 & \textbf{76.2} 
& 83.5 & 97.3 & 89.0 & 99.3 & 94.7 & \textbf{99.8} \\

\textbf{IMDB – Test} 
& 66.8 & \textbf{66.8} & 66.5 & \textbf{66.8} 
& 72.5 & \textbf{76.2} & 73.3 & 74.8 & 74.8 & 75.5 \\

\bottomrule
\end{tabular}
}
\end{table}

\subsection{Full-Dataset Evaluation Results}
\label{appendix:full_dataset_results}

In this subsection, we report results obtained by running all prompting-based baselines, TF-IDF + CART and ACT on the full training datasets for the BANKCHURN, SPAM and DIAGNO tasks. 

\paragraph{Full Datasets.} Table~\ref{tab:full_datasets_results} shows performance for ACT and baseline methods on the full datasets. For BANKCHURN, the dataset contains 7'731 training samples (20.57\% labeled ``yes'') and 1'969 test samples (19.55\% ``yes''). A validation set of 300 samples was constructed by random sampling from the training set while preserving the class distribution. When using this validation split, the effective training size is 7'431 samples. For SPAM, the training set has 7'766 (50.46\% ``yes'') and the test set has 2'725 (50.46\% ``yes'') samples. The validation set of 409 samples was constructed by random sampling from the training set while preserving the class distribution. For DIAGNO, the full test set containing the Tuberculosis (TB) and Allergic sinusitis samples contains 997 (33.5\% TB, resp. ``yes'') samples and the train set contains the maximum possible amount of samples (1994 with 33.5\% TB, resp. ``yes') such that it matches the test set distribution. The original train set distribution would be 50\% of each class, but we are interested in evaluating the in-distribution test accuracy. In Table~\ref{tab:full_datasets_results} the TF-IDF + CART configuration with the best test set accuracy among depths $3-10$ is reported and additionally Table~\ref{tab:tfidf_full_reference} reports TF-IDF + CART results across all tested tree depths.

% =================== Full Datasets TABLE =============================

\begin{table}[htbp!]
\centering
\small
\caption{Results for ACT and baselines (CoT, TestGrad, DSPY and TF-IDF+CART) methods on the full datasets for BANKCHURN, SPAM and DIAGNO. Metrics are test set accuracy and F1-score on the test set. Depending on the class imbalanced, different ACT hyperparameter configurations ($d$ for maximum depth and $k$ for number of optimization steps per node) were evaluated.}
\label{tab:full_datasets_results}

\resizebox{0.95\textwidth}{!}{
\begin{tabular}{p{2.2cm} l l c c c}
\toprule
\textbf{Dataset} & \textbf{Model} & \textbf{Method} & \textbf{Train Acc} & \textbf{Test Acc} & \textbf{Test F1} \\
\midrule

% ================= BANKCHURN =================

{\textbf{BANKCHURN}} & Gemma 
& CoT & 48.5 & 47.3 & 27.3 \\
& & TextGrad & 78.0 & 78.9 & 2.4 \\
& & DSPy & 55.5 & 56.0 & 33.2 \\
& & ACT $(d=4, k=20)$ & 75.9 & 75.0 & 22.9 \\
& & ACT $(d=8, k=10)$ & 82.6 & \textbf{82.7} & \textbf{40.6} \\
\noalign{\vskip 1.5pt}
\cdashline{2-6}
\noalign{\vskip 1.5pt}
& Nano 
& CoT & 66.7 & 67.8 & 19.1 \\
& & TextGrad & 79.4 & 80.5 & 0.0 \\
& & DSPy & 79.4 & 80.3 & 0.0 \\
& & ACT $(d=4, k=20)$ & 80.2 & 80.7 & 9.1 \\
& & ACT $(d=8, k=10)$ & 80.7 & \textbf{80.7} & \textbf{12.4} \\
\noalign{\vskip 1.5pt}
\cdashline{2-6}
\noalign{\vskip 1.5pt}
& $-$ & TF-IDF + CART (best, $d{=}3$) 
& 79.7 & 80.7 & 4.5 \\

\midrule

% ================= SPAM =================

\textbf{SPAM} & Gemma 
& CoT & 74.9 & 75.0 & 80.1 \\
& & TextGrad & 97.5 & 97.2 & 97.2 \\
& & DSPy & 97.1 & 97.4 & 97.5 \\
& & ACT $(d=2, k=10)$ & 96.5 & 96.3 & 96.4 \\
& & ACT $(d=3, k=10)$ & 97.9 & 97.6 & 97.6 \\
& & ACT $(d=4, k=10)$ & 98.7 & 97.9 & 98.0 \\
& & ACT $(d=5, k=10)$ & 99.1 & \textbf{98.8} & \textbf{98.8} \\
\noalign{\vskip 1.5pt}
\cdashline{2-6}
\noalign{\vskip 1.5pt}
& Nano 
& CoT & 96.1 & 95.6 & 95.5 \\
& & TextGrad & 98.1 & 98.2 & 98.2 \\
& & DSPy & 98.1 & 98.2 & 98.2 \\
& & ACT $(d=2, k=10)$ & 99.6 & 99.5 & 99.5 \\
& & ACT $(d=3, k=10)$ & 99.6 & 99.5 & 99.5 \\
& & ACT $(d=4, k=10)$ & 99.7 & \textbf{99.6} & \textbf{99.6} \\
& & ACT $(d=5, k=10)$ & 99.6 & 99.5 & 99.5 \\
\noalign{\vskip 1.5pt}
\cdashline{2-6}
\noalign{\vskip 1.5pt}
& $-$ 
& TF-IDF + CART (best, $d{=}10$) 
& 95.3 & 94.9 & 95.2 \\

\midrule
% ================= DIAGNO =================

\textbf{DIAGNO} & Gemma 
& CoT & 49.2 & 50.1 & 55.0 \\
& & TextGrad & 68.6 & 67.0 & 54.2 \\
& & DSPy & 67.6 & 66.5 & 57.9 \\
& & ACT $(d=4, k=10)$ & 73.6 & 72.6 & 41.5 \\
& & ACT $(d=5, k=10)$ & 74.8 & \textbf{72.8} & 52.7 \\
& & ACT $(d=8, k=10)$ & 81.4 & 71.9 & \textbf{59.4} \\
\noalign{\vskip 1.5pt}
\cdashline{2-6}
\noalign{\vskip 1.5pt}
& Nano 
& CoT & 69.0 & 68.6 & 48.1 \\
& & TextGrad & 69.4 & 68.6 & 48.1 \\
& & DSPy & 73.8 & 70.7 & 45.9 \\
& & ACT $(d=4, k=10)$ & 74.9 & 72.7 & 54.1 \\
& & ACT $(d=5, k=10)$ & 75.8 & \textbf{73.7} & 54.2 \\
& & ACT $(d=8, k=10)$ & 80.2 & 73.4 & \textbf{56.1} \\
\noalign{\vskip 1.5pt}
\cdashline{2-6}
\noalign{\vskip 1.5pt}
& $--$ 
& TF-IDF + CART (best, $d{=}7$) 
& 90.1 & 85.5 & 76.7 \\

\bottomrule
\end{tabular}
}
\end{table}

\paragraph{Results.} On the heavily imbalanced BANKCHURN dataset, several baselines (e.g., TextGrad and DSPy with GPT-4.1-Nano) collapse to majority-class prediction, yielding F1-scores of 0, while ACT with Gemma-3-4b achieves the highest accuracy among all methods alongside an acceptable F1-score. On SPAM, where the label distribution mirrors the random subset used in the main paper, prompting-based methods achieve results comparable to Table~\ref{tab:results_acart}, confirming the effectiveness of prompt-based methods in limited data settings; TF-IDF+CART improves modestly with the full dataset but remains below the prompting-based baselines. On DIAGNO, TF-IDF+CART outperforms prompt-based baselines built on small LLMs, consistent with the random subset results in Table~\ref{tab:results_acart}; however, additional results on ACT with GPT-4.1-Mini $(d=8, s=10)$ achieves 81.4\% accuracy and an F1-score of 68.4, demonstrating that stronger backbone models close this gap. Crucially, unlike black-box prompt-based baselines or TF-IDF+CART, ACT produces interpretable and auditable decision paths. More broadly, larger training sets allow ACT to benefit from deeper trees without the overfitting observed on small subsets and performance differences between similarly-sized LLMs (Gemma-3-4b vs. GPT-4.1-Nano) become more pronounced on full datasets, particularly in F1-score on imbalanced tasks. Table~\ref{tab:tfidf_full_reference} shows that increasing tree depth consistently improves training accuracy, yet TF-IDF+CART overfits on BANKCHURN and DIAGNO. On the heavily imbalanced BANKCHURN task, the best test accuracy is achieved at depth 3 but at the cost of a near-zero F1-score, while depth 10 recovers F1 at the expense of accuracy, highlighting the difficulty of class imbalance.

% =================== TF-IDF + CART Full Datasets TABLE =============================

\begin{table}[htbp!]
\centering
\small
\caption{TF-IDF + CART results across all evaluated tree depths on the full BANKCHURN, SPAM and DIAGNO datasets.}
\label{tab:tfidf_full_reference}

\resizebox{0.8\textwidth}{!}{
\begin{tabular}{lcccccc}
\toprule
\textbf{Dataset} (Train/Test Set Size) & \textbf{Depth} & \textbf{Train Acc (\%)} & \textbf{Test Acc (\%)} & \textbf{Test F1 (\%)} \\
\midrule

% -------------------- BANKCHURN --------------------
\textbf{BANKCHURN} (7731/1969) 
& 3  & 79.7 & \textbf{80.7} & 4.5 \\
& 4  & 80.0 & 79.8 & 3.4 \\
& 5  & 80.4 & 80.2 & 4.4 \\
& 6  & 80.8 & 80.0 & 5.7 \\
& 7  & 81.3 & 79.4 & 8.6 \\
& 8  & 82.1 & 78.4 & 10.2 \\
& 9  & 82.7 & 79.6 & 10.7 \\
& 10 & \textbf{83.3} & 77.6 & \textbf{11.6} \\

\midrule

% -------------------- SPAM --------------------
\textbf{SPAM} (7766/2725)
& 3  & 90.9 & 90.2 & 91.2 \\
& 4  & 92.4 & 92.0 & 92.6 \\
& 5  & 93.5 & 93.4 & 93.9 \\
& 6  & 94.3 & 94.2 & 94.6 \\
& 7  & 94.9 & 94.8 & 95.1 \\
& 8  & 95.3 & 94.9 & 95.1 \\
& 9  & 95.3 & 94.9 & 95.2 \\
& 10 & \textbf{95.3} & \textbf{94.9} & \textbf{95.2} \\

\midrule

% -------------------- DIAGNO --------------------
\textbf{DIAGNO} (1994/997)
& 3  & 82.4 & 80.0 & 71.0 \\
& 4  & 83.2 & 80.9 & 72.6 \\
& 5  & 84.9 & 81.6 & 68.3 \\
& 6  & 88.0 & 84.6 & 75.6 \\
& 7  & 90.1 & \textbf{85.5} & \textbf{76.7} \\
& 8  & 91.7 & 83.9 & 74.3 \\
& 9  & 92.7 & 83.6 & 73.3 \\
& 10 & \textbf{94.1} & 83.5 & 73.9 \\

\bottomrule
\end{tabular}
}
\end{table}

\newpage

\section{Interpretability Statistics for ACT}
\label{appendix:interpretability}

In Table~\ref{tab:interpret_all} we report two interpretability indicators for ACT across all datasets and ACT configurations evaluated: the average question length (in characters) and the average path length (number of nodes traversed during inference). These values are averaged over the four LLMs used in our experiments. Across datasets, average question lengths range from 78 to 155 characters, reflecting concise node-level prompts, while average path lengths show that predictions typically require only about three nodes, even for depth-4 trees. We also observe a mild trade-off in which datasets with longer questions (e.g., JAILBREAK and SPAM) yield shorter paths, indicating that ACT can consolidate multiple relevant features into a single informative query. For reference, the generic seed question \textit{“Based on the provided example, does it belong to the positive class? (yes/no)”} has a length of 87 characters.

\begin{table}[h!]
\centering
\caption{Interpretability statistics across datasets.}
\small

% ==== Row 1 ====
\begin{subtable}[t]{0.45\textwidth}
\centering
\begin{tabular}{lcc}
\toprule
\textbf{ACT (d, k)} & \textbf{Avg. Q. Len} & \textbf{Avg. Path} \\
\midrule
(3, 10) & 92.00 & 2.35 \\
(3, 20) & 93.15 & 2.51 \\
(4, 10) & 78.05 & 3.03 \\
(4, 20) & 80.53 & 3.06 \\
\bottomrule
\end{tabular}
\caption{DIAGNO}
\label{tab:interpretdiagno}
\end{subtable}
\hfill
\begin{subtable}[t]{0.45\textwidth}
\centering
\begin{tabular}{lcc}
\toprule
\textbf{ACT (d, k)} & \textbf{Avg. Q. Len} & \textbf{Avg. Path} \\
\midrule
(2, 5)  & 118.00 & 1.33 \\
(2, 10) & 142.88 & 1.26 \\
(3, 5)  & 94.18 & 1.36 \\
(3, 10) & 122.63 & 1.14 \\
\bottomrule
\end{tabular}
\caption{SPAM}
\label{tab:interpretspam}
\end{subtable}

\vspace{0.5cm}

% ==== Row 2 ====
\begin{subtable}[t]{0.45\textwidth}
\centering
\begin{tabular}{lcc}
\toprule
\textbf{ACT (d, k)} & \textbf{Avg. Q. Len} & \textbf{Avg. Path} \\
\midrule
(3, 10) & 136.00 & 1.98 \\
(3, 20) & 155.00 & 2.13 \\
(4, 10) & 142.00 & 2.41 \\
(4, 20) & 150.33 & 2.57 \\
\bottomrule
\end{tabular}
\caption{JAILBREAK}
\label{tab:interpretjailbreak}
\end{subtable}
\hfill
\begin{subtable}[t]{0.45\textwidth}
\centering
\begin{tabular}{lcc}
\toprule
\textbf{ACT (d, k)} & \textbf{Avg. Q. Len} & \textbf{Avg. Path} \\
\midrule
(3, 10) & 90.50 & 2.92 \\
(3, 20) & 89.90 & 2.93 \\
(4, 10) & 82.20 & 3.10 \\
(4, 20) & 89.77 & 2.96 \\
\bottomrule
\end{tabular}
\caption{BANKCHURN}
\label{tab:interpretbankchurn}
\end{subtable}

\vspace{0.5cm}

% ==== Row 3 (centered single table) ====
\begin{subtable}[t]{0.6\textwidth}
\centering
\begin{tabular}{lcc}
\toprule
\textbf{ACT (d, k)} & \textbf{Avg. Q. Len} & \textbf{Avg. Path} \\
\midrule
(3, 10) & 81.30 & 1.97 \\
(3, 20) & 95.80 & 2.28 \\
\bottomrule
\end{tabular}
\caption{IMDB}
\label{tab:interpretimdb}
\end{subtable}
\label{tab:interpret_all}
\end{table}

\section{Additional Ablation Studies and Scaling Analysis}
\label{appendix:add_ablations}

\subsection{Ablation on the Number of Feedback Samples $m$}

We conduct an ablation study on the number of labeled samples $m$ used to generate semantic feedback during node-level prompt optimization. We evaluate $m \in \{5, 10, 20, 50\}$ on the DIAGNO dataset using GPT-4.1-nano for two ACT configurations: $(d=3, k=10)$ and $(d=4, k=20)$. The results in Tables~\ref{tab:mablation_d3} and \ref{tab:mablation_d4} show how varying $m$ affects training performance, test metrics, and token usage. As comparison, the results for DSPy are also shown.

\begin{table}[h!]
\centering
\caption{Effect of varying $m$ for ACT $(d=3, k=10)$ on DIAGNO with GPT-4.1-nano.}
\small
\begin{tabular}{lcccc}
\toprule
\textbf{m} & \textbf{Train Acc $\uparrow$} & \textbf{Test Acc $\uparrow$} & \textbf{Test F1 $\uparrow$} & \textbf{Tokens ($\times 100k$) $\downarrow$} \\
\midrule
5   & 69.8 & 67.0 & 67.6 & \textbf{52.4} \\
10  & 71.2 & 67.5 & 71.3 & 53.5 \\
20  & 71.3 & \textbf{69.5} & \textbf{71.7} & 54.2 \\
50  & 71.8 & 67.2 & 69.0 & 55.2 \\
DSPy & 71.5 & 68.7 & 61.2 & 122.1 \\
\bottomrule
\end{tabular}
\label{tab:mablation_d3}
\end{table}

\begin{table}[h!]
\centering
\caption{Effect of varying $m$ for ACT $(d=4, k=20)$ on DIAGNO with GPT-4.1-nano.}
\small
\begin{tabular}{lcccc}
\toprule
\textbf{m} & \textbf{Train Acc $\uparrow$} & \textbf{Test Acc $\uparrow$} & \textbf{Test F1 $\uparrow$} & \textbf{Tokens ($\times100k$) $\downarrow$} \\
\midrule
5   & 71.7 & 65.8 & 65.9 & 124.7 \\
10  & 74.2 & 68.7 & 70.6 & 135.0 \\
20  & 74.5 & 70.0 & \textbf{71.3} & 137.6 \\
50  & 74.8 & \textbf{70.3} & 70.8 & 143.8 \\
DSPy & 71.5 & 68.7 & 61.2 & \textbf{122.1} \\
\bottomrule
\end{tabular}
\label{tab:mablation_d4}
\end{table}

\subsection{Model Scaling and Cost-Normalized Analysis}
\label{appendix:model_scaling}

To assess how model strength affects different prompting strategies, we conducted an additional scaling study using GPT-4.1. We evaluate CoT, DSPy, TextGrad, and two ACT configurations on the DIAGNO and BANKCHURN datasets, as these are the most challenging tasks. As shown in Table~\ref{tab:model-scaling}, stronger models do not consistently improve CoT, DSPy, or TextGrad, whereas ACT benefits from increased model capacity. These results indicate that ACT more effectively leverages the enhanced reasoning abilities of more powerful models.

\begin{table}[h!]
\centering
\caption{Performance scaling of CoT, DSPy, TextGrad, and ACT when moving from GPT-4.1-nano to GPT-4.1-mini to GPT-4.1. Values shown are test-set accuracies in \%. ACT benefits more consistently from stronger models.}
\small
\begin{tabular}{lccc}
\toprule
\textbf{Method} & \textbf{GPT-4.1-nano} & \textbf{GPT-4.1-mini} & \textbf{GPT-4.1} \\
\midrule
\multicolumn{4}{c}{\textbf{DIAGNO}} \\
\midrule
CoT            & 63.3 & 61.2 & 62.2 \\
DSPy           & \textbf{68.7} & 64.5 & 68.2 \\
TextGrad       & 64.5 & 65.8 & 64.8 \\
ACT (3, 10)    & 66.9 & \textbf{77.3} & 80.2 \\
ACT (3, 20)    & 67.2 & 76.2 & \textbf{81.8} \\
\midrule
\multicolumn{4}{c}{\textbf{BANKCHURN}} \\
\midrule
CoT            & 48.5 & 52.8 & 57.2 \\
DSPy           & 52.3 & 54.0 & 59.5 \\
TextGrad       & 53.3 & 55.3 & 58.2 \\
ACT (3, 10)    & 58.8 & 60.5 & 69.2 \\
ACT (3, 20)    & \textbf{67.0} & \textbf{69.2} & \textbf{72.5} \\
\bottomrule
\end{tabular}
\label{tab:model-scaling}
\end{table}

Figure~\ref{fig:ablation_diagno_cost} in the main paper visualizes the accuracy–cost trade-off across methods and model sizes, highlighting how performance scales with increasing computational expense. This section further reports token usage, monetary cost, test accuracy, and cost-normalized accuracy for each method, with Table~\ref{tab:cost_norm} summarizing these metrics across the GPT-4.1 family on DIAGNO to facilitate comparison of efficiency and scalability.

Table~\ref{tab:cost_norm} shows that TextGrad achieves the highest cost-normalized accuracy in the Nano and Mini settings, while ACT (3,10) remains competitive—ranking just behind TextGrad in smaller models and performing strongly on GPT-4.1. The (3,20) variant is less cost-efficient due to higher token usage, though it still slightly outperforms DSPy on Mini and GPT-4.1. As shown in Figure~\ref{fig:ablation_diagno_cost} in the main paper, ACT’s test accuracy consistently scales with cost across all model sizes, whereas TextGrad and DSPy do not exhibit consistent cost-dependent accuracy gains.

% \begin{figure}[hb!]
%     \centering
%     \includegraphics[width=\textwidth]{images/diagno_cost_norm.pdf}
%     \caption{\textbf{Accuracy–cost trade-off across methods and model scales.}
%     The plot shows test accuracy versus monetary cost for ACT, TextGrad, and DSPy across the Nano, Mini, and GPT-4.1 model families. Each method is connected across model scales to illustrate how performance changes with increasing cost.}
%     \label{fig:cost_diagno}
% \end{figure}

\begin{table}[ht!]
\centering
\caption{Cost normalization and model scaling across the GPT-4.1 family on the DIAGNO dataset.}
\label{tab:cost_norm}
\begin{tabular}{lcccccc}
\toprule
\textbf{Methods} & \textbf{Model} & \textbf{Cost $\downarrow$} & \textbf{Test Acc. $\uparrow$} & \textbf{Total Tokens (x100) $\downarrow$} & \textbf{Cost Norm Acc. $\uparrow$} \\
\midrule
ACT (3, 10)    & Nano    & 1.09  & 66.9 & 55.18  & 61.54 \\
ACT (3, 20)    & Nano    & 2.02  & 67.2 & 101.37 & 33.31 \\
DSPY     & Nano    & 1.85  & \textbf{68.7} & 152.58 & 37.23 \\
TextGrad & Nano    & \textbf{0.80}  & 64.5 & \textbf{48.06}  & \textbf{80.93} \\
\midrule
ACT (3, 10)    & Mini    & 3.09  & \textbf{77.3} & 48.08  & 24.98 \\
ACT (3, 20)    & Mini    & 5.58  & 76.2 & 87.58  & 13.65 \\
DSPY     & Mini    & 7.92  & 64.5 & 161.28 & 8.15  \\
TextGrad & Mini    & \textbf{2.30}  & 65.8 & \textbf{36.37}  & \textbf{28.60} \\
\midrule
ACT (3, 10)    & GPT-4.1 & 19.08 & 80.2 & 53.71  & \textbf{4.20}  \\
ACT (3, 20)    & GPT-4.1 & 36.11 & \textbf{81.8} & 102.46 & 2.26  \\
DSPY     & GPT-4.1 & 37.10 & 68.2 & 153.89 & 1.84  \\
TextGrad & GPT-4.1 & \textbf{17.76} & 64.8 & \textbf{49.08}  & 3.65  \\
\bottomrule
\end{tabular}
\end{table}

\section{Use of Large Language Models}

During the preparation of this manuscript, we made limited use of large language models to enhance the clarity and readability of the text. This involved assistance with grammar, phrasing, and stylistic improvements, particularly in the abstract and selected explanatory sections. All scientific content, including the formulation of research questions, experimental design, results, and interpretations, was developed solely by the authors. No language model was used to generate original ideas, proofs, or analyses.

\end{document}